\documentclass{article}
\usepackage[utf8]{inputenc}

\usepackage{amsmath}
\usepackage{amssymb}
\usepackage{amsthm}
\usepackage[english]{babel}
\usepackage{bbold}
\usepackage{charter}
\usepackage{fullpage}
\usepackage{hyperref}
\usepackage{pifont}
\usepackage{color,soul}
\usepackage{listings}
\usepackage{lipsum}
\usepackage{empheq}
\usepackage{subfigure}
\usepackage[most]{tcolorbox}
\usepackage{graphicx}
\usepackage{multirow}
\usepackage{xcolor}
\usepackage{tabularx}
\usepackage{makecell}
\usepackage{xspace} 

\newtcbox{\mymath}[1][]{%
    nobeforeafter, math upper, tcbox raise base,
    enhanced, colframe=gray!15!black,
    colback=gray!15, boxrule=0.5pt,
    #1}
    
\newcommand\upload{\textcolor{blue}{\uparrow}}
\newcommand\download{\textcolor{red}{\downarrow}}
\newcommand{\sys}{\textsc{Bagua}\xspace}
\newcommand{\cmark}{\ding{51}}%

\definecolor{codegreen}{rgb}{0,0.6,0}
\definecolor{codegray}{rgb}{0.5,0.5,0.5}
\definecolor{codepurple}{rgb}{0.58,0,0.82}
\definecolor{backcolour}{rgb}{0.95,0.95,0.92}
\newcommand{\blue}[1]{\textcolor{black}{#1}}

\title{\bf \sys: Scaling up Distributed Learning \\with System Relaxations}

\author{
    Shaoduo Gan$^*$, Jiawei Jiang,\\Binhang Yuan, Ce Zhang\\
    \normalsize ETH Z\"urich, Switzerland\\
    \normalsize \{sgan, jiawei.jiang, \\
    \normalsize binhang.yuan, ce.zhang\}@inf.ethz.ch \\

  \and
    Xiangru Lian$^*$, Rui Wang, Jianbin Chang,\\ Chengjun Liu, Hongmei Shi, Shengzhuo Zhang,\\ Xianghong Li, Tengxu Sun, Sen Yang, Ji Liu\\
    \normalsize Kuaishou Technology, China\\
    \normalsize admin@mail.xrlian.com\\
    \normalsize ji.liu.uwisc@gmail.com\\
}

\date{}

\begin{document}
\maketitle
\begin{abstract}
\blue{Recent years} have witnessed a growing list of systems for distributed data-parallel training. Existing systems largely fit into two paradigms, i.e., parameter server and MPI-style collective operations. On the algorithmic side, researchers have proposed a wide range of techniques to lower the communication via ``system relaxations'': \textit{quantization}, \textit{decentralization}, and \textit{communication delay}. However, most, if not all, existing systems only rely on standard synchronous and asynchronous stochastic gradient (SG) based optimization, therefore, cannot take advantage of all possible optimizations that the machine learning community has been developing recently. Given this emerging gap between the current landscapes of systems and theory, we build \sys, a {MPI-style} communication {library}, {providing a collection of primitives}, that is both flexible and modular to support state-of-the-art system relaxation techniques of distributed training. Powered by this design, \sys has a great ability to implement and extend various state-of-the-art distributed learning algorithms. In a production cluster with up to 16 machines (128 GPUs), $\sys$ can outperform PyTorch-DDP, Horovod and BytePS in the end-to-end training time by a significant margin (up to 2$\times$) across a diverse range of tasks. Moreover, we conduct a rigorous tradeoff exploration showing that different algorithms and system relaxations achieve the best performance over different network conditions. 
\end{abstract}

\begingroup
\renewcommand\thefootnote{}
\footnote{\noindent
* Equal contribution.
}
\footnote{\noindent
\sys is publicly available at \url{https://github.com/BaguaSys/bagua}.
}
\addtocounter{footnote}{-1}
\endgroup

\section{Introduction}

The increasing scalability and performance 
of distributed machine learning systems has been one of the main
driving forces behind the rapid 
advancement of machine learning techniques. 
From AlexNet~\cite{krizhevsky2012imagenet} in 2012 to 
GPT-3~\cite{brown2020language} in 2020, each
leap in model quality is enabled by the growth of both the model size and
the amount of data one can train a model with,
along with a rapid increase in 
computations~\cite{mayer2020scalable}. Behind this improvement
are two major enabling factors: 
hardware accelerations
(e.g., GPUs and TPUs) and
the development of efficient and scalable 
distributed training algorithms
~\cite{alistarh2016qsgd,zhang2017zipml,bernstein2018signsgd,wen2017terngrad,wangni2018gradient}
It is not unfair to say that a scalable distributed
training system is the cornerstone 
of modern deep learning techniques.

\paragraph*{\underline{Current Landscape
of Data Parallel Training Systems}}
\blue{In this paper, we scope 
ourselves and focus on \textit{data parallel
training}, one of the most popular 
distributed training paradigms
in which the data set is partitioned across different workers and the model 
fits into a single device.} Not surprisingly, recently years have 
witnessed a growing list of systems
for distributed data parallel training. Existing systems 
fit into two paradigms, following the 
seminal work done by Li et al.~\cite{li2014scaling}
on \textit{parameter server} and Sergeev et al.~\cite{sergeev2018horovod} on using MPI collective operations such as \textit{Allreduce}. 
Both paradigms have enabled industrial-scale distributed training systems~\cite{mayer2020scalable}: Adam (Microsoft)~\cite{chilimbi2014project}, early TensorFlow (Google)~\cite{abadi2016tensorflow}, Poseidon (Petuum)~\cite{zhang2017poseidon}, Angel (Tencent)~\cite{jiang2018angel}, and BytePS (ByteDance)~\cite{jiang2020unified} are based on \textit{parameter server}, while PyTorch-DDP (Facebook)~\cite{li13pytorch}, Mariana (Tencent)~\cite{zou2014mariana}, MALT (NEC Labs)~\cite{li2015malt}, NCCL (NVIDIA)~\cite{nccl}, and Horovod (Uber)~\cite{sergeev2018horovod} are based on \textit{MPI-style collective operations}.
These systems often involve 
joint efforts from machine learning, systems, and data management communities, and have been 
successful in making distributed training 
easier and more scalable.

\paragraph*{\underline{Current Landscape
of Data Parallel Training Algorithms}}
\blue{On the theory and algorithm side,
researchers have also been active in 
improving the performance of 
standard \textit{synchronous} and \textit{asynchronous} 
stochastic gradient (SG) based
algorithms.} Rightly noticing that 
a major system bottleneck is  
communication, researchers have 
proposed a range of techniques 
to lower the communication overhead
mainly by ``\textit{relaxing}'' certain aspects
of the communication.
Examples include (1) \textit{communication compression} (e.g., 
quantization~\cite{alistarh2016qsgd,zhang2017zipml,bernstein2018signsgd,wen2017terngrad}, 
sparsification~\cite{wangni2018gradient,alistarh2018convergence,wang2018atomo,wang2017efficient}, and 
error compensation~\cite{tang2019doublesqueeze}),
(2) \textit{communication decentralization}~\cite{koloskova2019decentralized,li2018pipe,lian2017can,lian2018asynchronous,tang2018communication,tang2018d}, and (3) \textit{communication delay} (e.g., LocalSGD~\cite{wang2019adaptive,lin2019don,stich2018local,haddadpour2019local}) and \textit{asynchronization}~\cite{lian2018asynchronous, zhou2018distributed, simsekli2018asynchronous, zheng2017asynchronous,peng2017asynchronous}.
These techniques are optimized for 
\textit{different workloads}
and different \textit{network conditions}.
These techniques \textit{together} hold 
promises 
to significantly decrease the 
communication overheads, in terms of both
bandwidth and latency, or increase the 
tolerance to the existence of stragglers.

\paragraph*{\underline{An Emerging Gap between System and Theory}}
\blue{In this paper, we are motivated by one 
emerging gap between the current landscapes 
of systems and theory:}
\textit{Despite the recent advance of 
distributed learning theory and algorithm
on system relaxations,
most, if not all, existing systems only rely on standard synchronous and asynchronous stochastic gradient (SG) based algorithms.} The main 
consequence is that existing systems 
are not taking advantage of all possible 
optimizations that the machine learning
community has been developing, and potentially 
many real-world applications can be 
further accelerated. 
In this paper, we ask: 
\textit{Can we further accelerate 
distributed learning systems 
with system relaxations for communications? 
If so, what is the right abstraction
for this purpose?}

\begin{table}[t!]
\centering
\begin{tabular}{@{}c@{\hspace{0.5em}}c@{\hspace{0.5em}}c@{\hspace{0.5em}}c@{\hspace{0.5em}}|@{\hspace{0.5em}}c@{\hspace{0.5em}}c@{\hspace{0.5em}}c@{\hspace{0.5em}}c@{}}
\hline
{\bf Alg.} & {\bf Sync.} & {\bf Precision} & {\bf Centralization} & {\bf PyTorch-DDP} & {\bf Horovod} & {\bf BytePS} & {\bf \sys}  \\
\hline
\cite{li2014scaling} & Sync. & Full Prec. & Centralized & \cmark & \cmark & \cmark & \cmark \\ 
\cite{koloskova2019decentralized} & Sync. & Full Prec. & Decentralized & ~ & ~ & ~ & \cmark \\ 
\cite{alistarh2016qsgd,stich2018sparsified} & Sync. & Low Prec. & Centralized & \cmark & \cmark & \cmark & \cmark \\ 
\cite{tang2018communication,tang2018d} & Sync. & Low Prec. & Decentralized & ~ & ~ & ~ & \cmark \\ 
\cite{zheng2017asynchronous} & Async. & Full Prec. & Centralized & ~ & ~ & \cmark & \cmark$^{*}$ \\ 
\cite{lian2018asynchronous} & Async. & Full Prec. & Decentralized & ~ & ~ & ~ & \cmark$^{*}$ \\ 
\cite{de2017understanding} & Async. & Low Prec. & Centralized & ~ & ~ & ~ & \cmark$^{*}$ \\ 
- & Async. & Low Prec. & Decentralized & ~ & ~ & ~ & ~ \\ 
\hline
\end{tabular}
\caption{\blue{Different system relaxation techniques. Async algorithms let works communicate without waiting for the computation or other workers. Low precision algorithms use compression techniques to compress the communication data. Decentralized algorithms remove the requirement of collecting data globally. The goal of \sys is to support these diverse communication patterns.}}
\label{fig:landscape}
\end{table}

\paragraph*{\underline{The \sys System and Our Contributions}}
\blue{In this paper,  
we present \sys, a communication library 
whose goal is to support 
state-of-the-art system relaxation techniques 
of distributed training.} We made two technical 
contributions.

\blue{\textbf{Our first contribution} is the system design
of \sys, which provides a modular design 
for communications.}
\sys \blue{is a natural extension of 
the popular parameter server and Allreduce paradigms,
inspired by the challenges of directly adapting these
paradigms to support algorithms in Table~\ref{fig:landscape}
(See Section~\ref{sec:discussion} for details).
Specifically, we provide a collection of MPI-style collective operations to facilitate communication with different precision and centralization strategies.}
{These primitives are} 
flexible and modular enough to 
support many
algorithms, illustrated in Table~\ref{fig:landscape}.
Moreover, we also develop 
a simple \textit{automatic optimization framework} 
that speeds up algorithms
implemented within the \sys framework.
The key behind this framework is 
automatic batching and 
scheduling of communications. Different from 
previous work such as 
Horovod~\cite{sergeev2018horovod} and BytePS~\cite{jiang2020unified},
our optimization framework can be 
applied more widely 
beyond the standard SG based algorithm.

\blue{\textbf{Our second contribution} is an extensive
empirical study centered around two 
hypotheses: (1) \textit{By supporting 
different system relaxation techniques,
\sys is able to provide significant 
improvement for real-world applications
and workloads with real-world infrastructure 
over existing systems}; and (2) 
\textit{By supporting a diverse range of 
system relaxations, \sys is able to 
provide a scalable ML training
over a diverse network conditions to 
allow a user picking different
algorithms. 
}}
To this end, we conduct a large-scale 
empirical study with both benchmark 
tasks and real-world applications 
running at Kwai Inc. 
On a cluster with up to 16 machines 
(128 GPUs in total, aggregated 2 petaFLOPS with Tensor Cores) 
we consider various network conditions
following how V100 GPU machines (\texttt{p3.8xlarge}, \texttt{p3.16xlarge}, \texttt{p3dn.24xlarge}) are connected on AWS:
10Gbps, 25Gbps, and 100Gbps, with TCP/IP connections.
\sys 
outperforms BytePS~\cite{jiang2020unified}, Horovod~\cite{sergeev2018horovod}, and PyTorch-DDP~\cite{li13pytorch}
by a significant margin (up to 2$\times$ for 10Gbps
and up to 1.34$\times$ for 100Gbps) 
across a diverse range of tasks.
Moreover, we conduct a rigorous tradeoff
exploration showing that different 
algorithms and system relaxations 
achieve best performance over different 
network conditions. This illustrates 
the importance of providing this 
diverse cohort of algorithms to 
an end user. 

\paragraph*{\underline{Limitations and Moving Forward}}
\blue{There are several limitations of
the current \sys system and we hope 
our efforts in building \sys can 
help and inspire future research in these 
exciting directions.} \underline{First},
\sys does not provide a principled way
to help a user to \textit{automatically}
pick the most suitable system relaxations
to apply. 
One exciting direction, 
after \sys provides the support for 
all these algorithms, is to understand 
how to build a principled auto-tuning system.
\underline{Second}, 
currently \sys only focuses on data parallelism and
it is interesting future work to integrate other
techniques such as model parallelism (e.g. \cite{shazeer2018mesh, shoeybi2019megatron, jia2019beyond,wang2019supporting,narayanan2020memory,lepikhin2020gshard, jankov2019declarative,yuan2020tensor}) 
and pipeline parallelism (e.g.,~\cite{huang2019gpipe,narayanan2019pipedream, li2021terapipe,he2021pipetransformer})
and to understand the system abstractions.

\paragraph*{\underline{Outline}}
The rest of the paper is organized as follows. We start by a brief review of data parallel training and the optimization frameworks of existing 
systems in Section \ref{sec:pre}, acting as both the preliminaries and related work. We
discuss the design and optimization of \sys in Section \ref{sec:sys_design}.
We describe our experimental study in Section \ref{sec:eval} and conclude in Section \ref{sec:con}.

\section{Preliminaries and Related Work}
\label{sec:pre}

\sys is built on decades of research regarding distributed 
machine learning systems and algorithms. Plenty of them are from the database community~\cite{SystemML,TeraSQLML,HybridParallelVLDB,VerticaML,DB4ML}.
We now summarize related work and discuss some in details to 
provide backgrounds and contexts. 
\blue{We refer the reader to
Appendix~\ref{sec:appendix_related_work} for a more detailed overview of distributed learning systems and
\cite{liu2020distributed} for the rigorous theoretical analysis of different system relaxation algorithms.}

\begin{figure}[t!]
\centering
\includegraphics[width=0.6\textwidth]{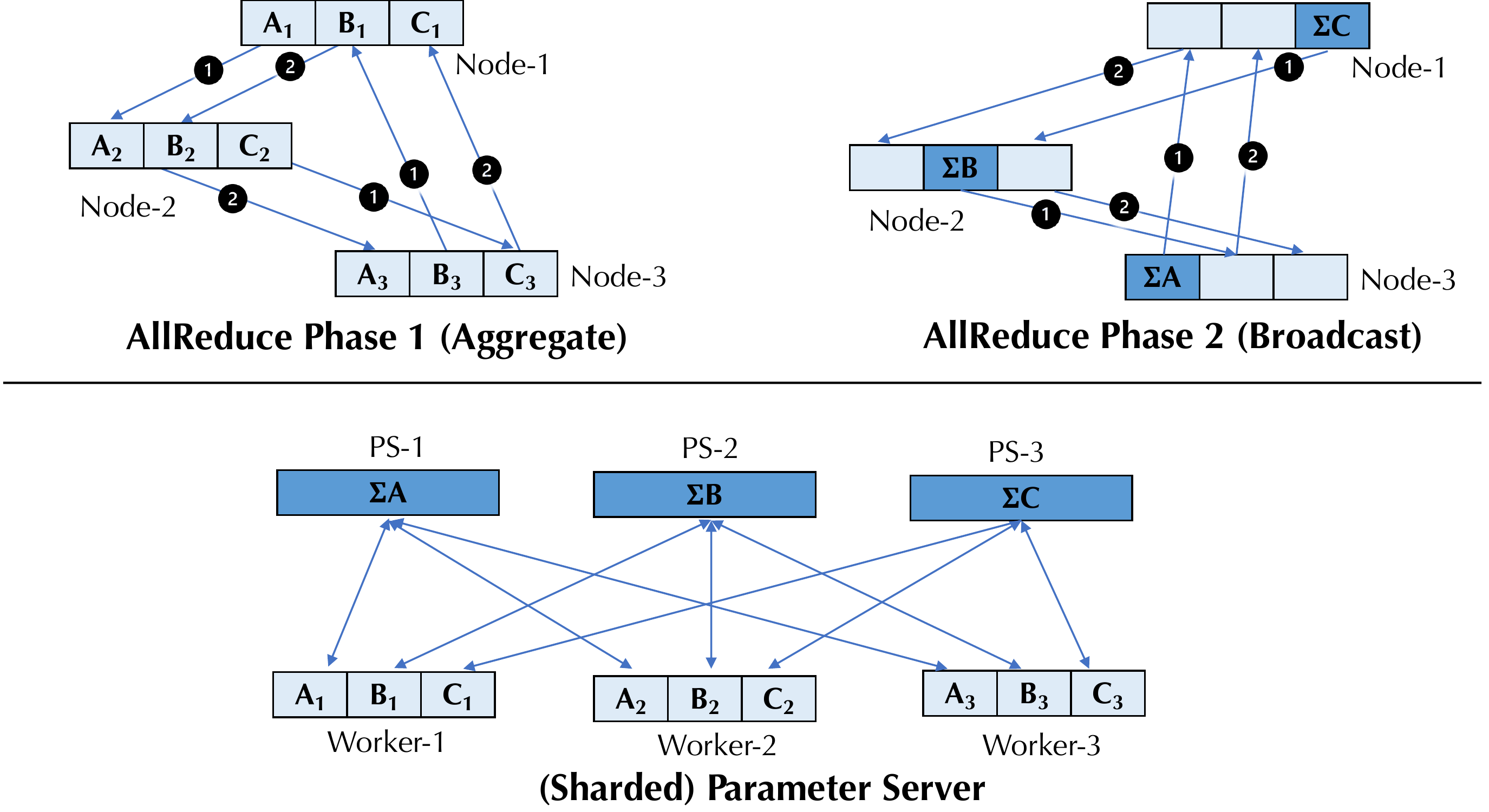}
\caption{Illustration of Parameter Server and \texttt{Allreduce}.}
\label{fig:ps_allreduce}
\end{figure}

\subsection{Data Parallel SG Based Algorithm}

The cornerstone of distributed learning systems is 
the data-parallel stochastic gradient based (DP-SG)
algorithms~\cite{li2014scaling}, which is the dominating algorithm that 
existing systems support and optimize for. 
Let $D$ be a dataset, 
$n$ is the number of workers, each worker $i$ holds its partition of
the data $D^{(i)}$ and model replica
at step $t$: $x_i^{(t)}$. Let $g_i^{(t)}$ be 
the stochastic gradient on worker $i$ at step $t$,
a textbook DP-SG updates each local model replica,
at worker $i$, as follows:%
{
\[
x^{(t+1)}_{i} = 
   x^{(t)}_{i} - \gamma \sum^n_{j=1} g^{(t)}_j
\]}
\noindent where $\gamma$ is the learning rate. To make this happen,
all machines need to exchange their local gradients
$g^{(t)}_i$, aggregate, and broadcast to all machines. 
Naturally, this can be implemented by the standard Allreduce
communication pattern.%

When there are many workers or some potential stragglers, 
one can extend the above algorithm into its 
asynchronous counterpart. 
Instead of using the latest 
gradient at iteration $t$, we allow the access to some staled 
version:%
{
\[
x^{(t+1)}_{i} = 
   x^{(t)}_{i} - \gamma \sum^n_{j=1} g^{(\tilde{t}_j^{(i)})}_{j}
\]}%
where $\tilde{t}_j^{(i)} \leq t$ is 
the previous iteration at which the gradient of 
worker $j$ is computed, accessed by the worker $i$ 
at iteration $t$. In theory, linear speedup can be achieved by async-SGD~\cite{liu2020distributed}.

\subsection{Existing Distributed Learning Systems}

Distributed learning systems have attracted intensive 
research over the last decade. 
Most existing systems, 
e.g., DistBelief~\cite{dean2012large}, Angel~\cite{jiang2018angel}, BytePS~\cite{jiang2020unified}, and
PyTorch-DDP~\cite{li13pytorch},
all focus on the optimization of the DP-SG
algorithm or its asynchronous counterpart. 
There are two fundamental questions governing the design of these systems:
\begin{enumerate}
\item \textit{(Abstraction for Communications) How should one communicate and aggregate the gradient and model?}
\item \textit{\textit{(Optimizations)} How should one optimize the end-to-end execution by balancing the communication and computation?}
\end{enumerate}

\blue{In terms of the abstraction for communications, existing systems
fall into two paradigms: parameter server (PS) \cite{dean2012large, li2014scaling, dai2015high,FlexpsVLDB,HeteroSIGMOD,PS2} and Allreduce \cite{sergeev2018horovod,zhang2019mllib,jiang2018dimboost,chen2016xgboost}.
Figure~\ref{fig:ps_allreduce} illustrates these two paradigms.} In a \textit{parameter server} architecture, the model can be partitioned to shards and distributed to multiple nodes (we call these nodes ``parameter servers''). 
During the training phase, workers periodically fetch the model from PS, leverage the computation unit like a GPU to conduct forward and backward propagations and push the gradients to the PS, while the PS aggregates the gradients and updates the parameters. 
With an \textit{Allreduce} paradigm, all the workers collaborate with their neighbors for model/gradient exchanges. A ring topology~\cite{rabenseifner2004optimization} is
often adopted by existing systems for a two-phase communication: first, the paradigm partitions the model/gradient into $n$ chunks (where $n$ is the number of nodes), and use $n$ rings with different starting and ending points to aggregate $n$ chunks; second, the aggregation result of each chunk located in different nodes is broadcast through the ring.

\begin{figure}[t!]
\centering
\includegraphics[width=0.7\textwidth]{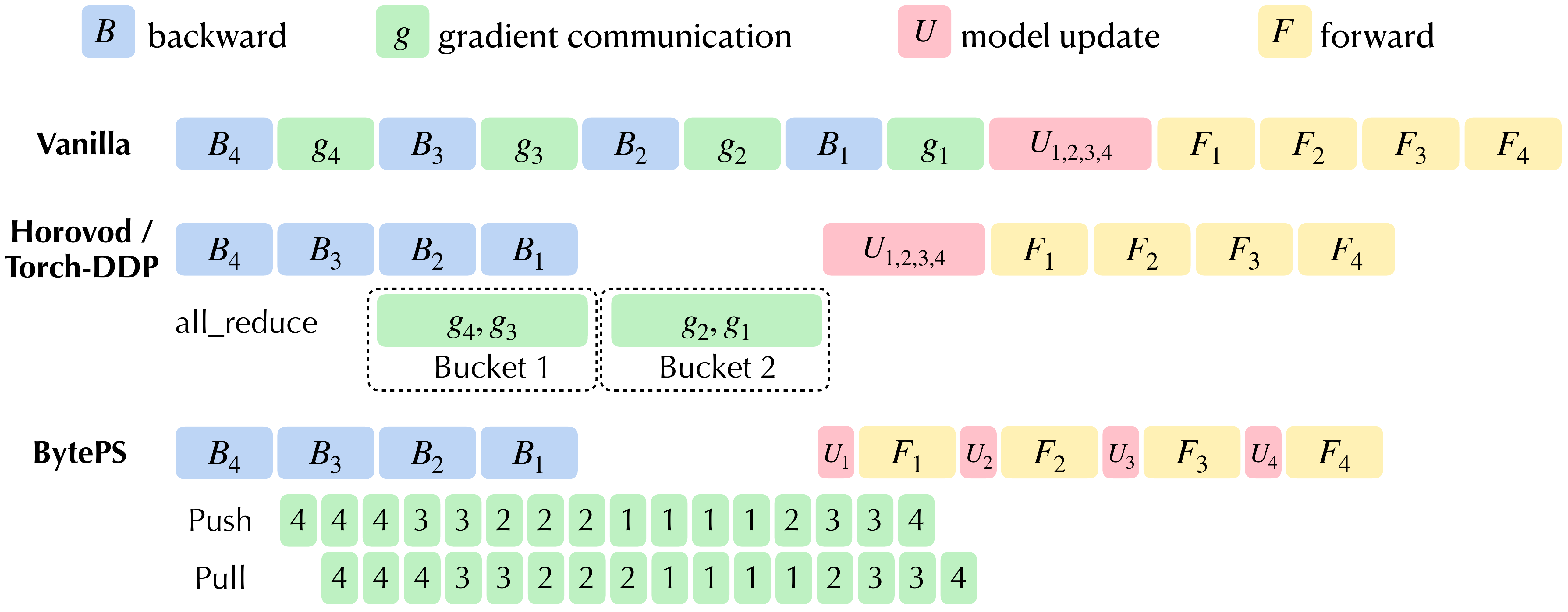}
\caption{Communication pattern of DP-SG and how Horovod, BytePS, and PyTorch-DDP optimizes the execution for this communication pattern.}
\label{fig:commpatterns}
\end{figure}

\blue{After deciding on which communication paradigm to use,
one key design is how to hide as much communication as possible during
computation.} This is often the core technical component
of previous systems, e.g., Horovod~\cite{sergeev2018horovod}, BytePS~\cite{jiang2020unified},
and PyTorch-DDP~\cite{li13pytorch}. 
These systems optimize the DP-SG communication pattern by developing different ways to balance communication and computation.
The key complexity roots from the fact that the training process of
DP-SG consists of delicate dependencies between different layers
and their own (1) forward pass, (2) backward pass, (3) gradient synchronization, and (4) model update, phases. Figure~\ref{fig:commpatterns} (Vanilla) illustrates a naive implementation of DP-SG over a model with four layers. The system would communicate 
gradient (green) for each layer 
once its backward pass (blue) finishes, and update the model for all layers (pink) in one go once all their 
communications are finished. The system then starts the next forward pass (yellow).

PyTorch-DDP and Horovod are two Allreduce-based systems, and have \textit{specifically} optimized this pipeline by overlapping the communication (Allreduce) with the backward pass and bucketing multiple gradients into one Allreduce operation. With overlapping, the Allreduce operations can take place in parallel with the computation of gradients. The Allreduce operation is only triggered when all gradients within a bucket are ready. The intuition of bucketing is that collective communications, like Allreduce, are more efficient on large tensors. After all Allreduce operations are finished, the model will be updated by the aggregated gradients.

\begin{figure}[t!]
\centering
\includegraphics[width=0.7\textwidth]{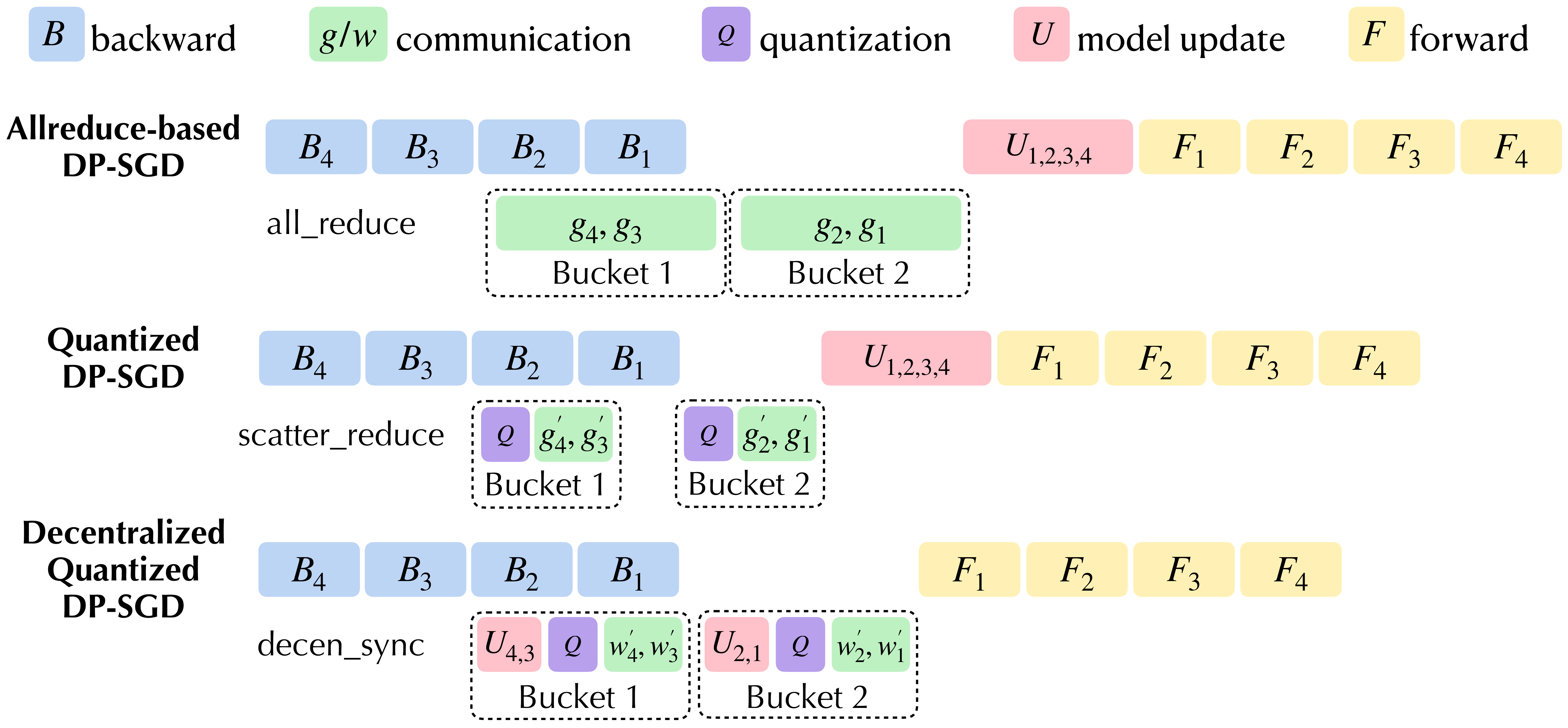}
\caption{\blue{Communication patterns of training algorithms with system relaxations, optimized by \sys automatically.}}
\label{fig:relaxation_patterns}
\end{figure}

BytePS, following the parameter server paradigm, has optimized this pipeline in a different way. BytePS partitions each gradient into small chunks with the identical size to conduct Push/Pull. BytePS overlaps Push/Pull with both
\textit{backward and forward} pass. It has a scheduler to maintain the communication order of gradient chunks. The principle is that parameters that are blocking the execution of the next forward pass will be prioritized for communication. Once all gradient chunks of a parameter have been pulled from the server, this parameter will be updated separately. Therefore, the forward pass of the next iteration could possibly be overlapped with the communication of the current iteration.
In terms of asynchronous DP-SG, BytePS supports
it by allowing each worker updating the state of the server individually without waiting for other workers. Whereas PyTorch-DDP and Horovod do not support asynchronous communications
since they rely on the Allreduce operator.

\subsection{System Relaxations for Distributed DP-SG}

While existing systems have been
mainly focusing on synchronous and asynchronous DP-SG algorithm,
the research community has developed a diverse set of techniques 
to further optimize for the different aspects of communications.
These techniques often lead to different training algorithms,
thus different communication patterns, as DP-SG. Given these
differences, none of Horovod, BytePS, and PyTorch-DDP provides
systematic support of these algorithms, as summarized in Table~\ref{fig:landscape}.
The goal of \sys is to provide a flexible abstraction to
support these diverse training algorithms with an automatic performance
optimization framework without assuming a specific communication pattern
such as the one of DP-SG.
Different strategies are proposed to speed up the expensive parameter exchange phase in DP-SG. In order to reduce communication volumes, lossy communication compression methods are introduced, such as quantization~\cite{alistarh2016qsgd,zhang2017zipml,bernstein2018signsgd,wen2017terngrad}, 
sparsification~\cite{wangni2018gradient,alistarh2018convergence,wang2018atomo,wang2017efficient}, sketching~\cite{ivkin2019communication}, and
error compensation~\cite{tang2019doublesqueeze}). 
In an attempt to get rid of the latency bottleneck, decentralized communication approaches are proposed~\cite{koloskova2019decentralized,li2018pipe,lian2017can,lian2018asynchronous,tang2018communication,tang2018d}. 
Additionally, localSGD is discussed to optimize for the number of communication rounds during training~\cite{wang2019adaptive,lin2019don,stich2018local,haddadpour2019local}.
To remove the synchronization barrier, which is an obstacle for clusters with very large number of workers and stragglers, some approach applies asynchronous update methods~\cite{peng2017asynchronous,zheng2017asynchronous,zhou2018distributed,simsekli2018asynchronous,nguyen2018sgd}. Lastly, it is worth to mention there are approaches that combines multiple strategies listed above~\cite{lian2018asynchronous,basu2019qsparse,koloskova2019decentralized,tang2019deepsqueeze,beznosikov2020biased}.

\blue{To illustrate the difference of communication patterns between these advanced training algorithms and vanilla DP-SG and the reason why systems that only have optimizing with DP-SG in mind faces challenges in supporting these new algorithms in a modular and systematic way, we take the example of QSGD~\cite{alistarh2016qsgd} and Decentralized Low-precision SGD ~\cite{tang2018communication}.} Figure~\ref{fig:relaxation_patterns} illustrates execution pipelines and communication patterns of DP-SG, QSGD and decentralized low-precision SGD. Compared with DP-SG, the execution components of the pipeline and their dependencies can be changed in the advanced algorithms. For example, the component "Quantization" required by both algorithms doesn't even exist in the DP-SG, and the "model update" component in Decentralized low-precision SGD needs to happen \textit{before} the communication. Since these advanced algorithms cannot fit into the DP-SG communication pattern, it is
challenging for systems born for DP-SG to handle 
these algorithms.

\section{System Design}
\label{sec:sys_design}

The goal of \sys is  
to support advanced training algorithms beyond DP-SG.
To achieve this, we revisit 
the two fundamental questions governing the design of 
previous systems, without assuming the pattern of DP-SG:
\begin{enumerate}
\item \textit{(Abstraction for Communications) How should one communicate and aggregate the gradient and model?} \textbf{In \sys, beyond parameter server and Allreduce, we design a collection of MPI-style collective operations to facilitate communications with different precision and centralization strategies.}
\item \textit{\textit{(Optimizations)} How should one optimize the end-to-end execution by balancing the communication and computation?} \textbf{In \sys, we develop a simple, but effective, automatic optimization framework which can be applied to optimize the execution of an algorithm implemented within \sys.}
\end{enumerate}
These two design decisions enable the flexibility and efficiency of \sys --- to implement a new advanced algorithm with system relaxation (e.g., 1-big Adam~\cite{tang20211} or Decentralized SGD~\cite{lian2017can}), in \sys, a developer does not need to worry about manually balancing communications with computations; instead,
she can specify, at a high-level, the logical semantics and 
\sys will automatically optimize its execution.
In this section, we first provide a high-level system 
overview, \blue{followed by the descriptions of these primitives} and their implementations, and then the simple, but effective, optimization framework in \sys.

\begin{figure}[t!]
\centering
\includegraphics[width=0.7\textwidth]{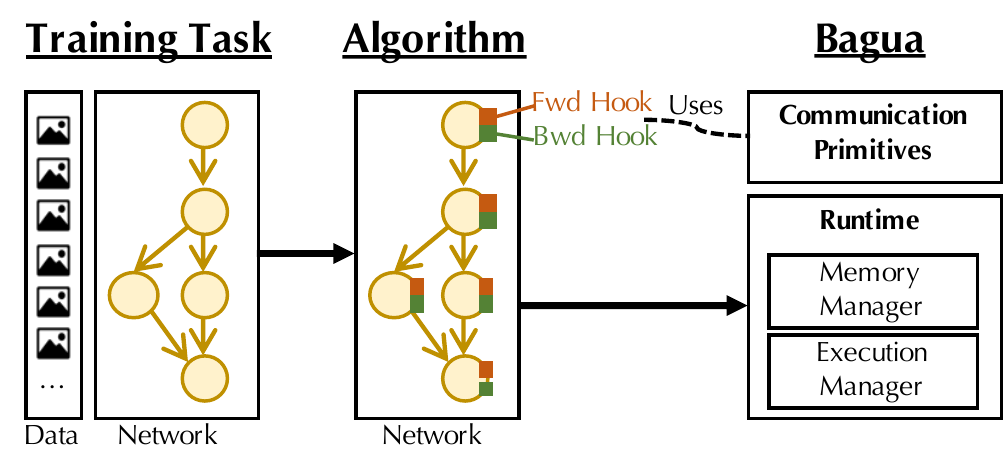}
\caption{Overview of \sys}
\label{fig:bagua}
\end{figure}

\subsection{System Overview}

The goal of \sys is to 
facilitate the development of efficient and scalable distributed training algorithms
that takes advantage of system
relaxations. As illustrated in Figure~\ref{fig:bagua}, there are 
three players: an \textit{end-user}, an \textit{optimization algorithm},
and the \textit{\sys runtime}.

\blue{From an end-user's perspective,} using 
\sys is very similar to 
use as PyTorch or TensorFlow for 
training on a single machine,
with minimal changes 
to their existing code.
\blue{The end-user should provide}: (1) a neural 
network model that needs
to train, specified as a 
graph in PyTorch,
and (2) a stream of data
examples. The end-user then 
specifies the training algorithm to use, e.g.,
QSGD~\cite{alistarh2016qsgd} (training with 
communication compression),
1-bit Adam~\cite{tang20211},
or DecentralizedSGD~\cite{lian2017can},
together with the information 
of the training infrastructure
such as the number of machines 
and whether one should use 
MPI or NCCL for communication.
\blue{We provide an example in Appendix~\ref{sec:program_train} to illustrate how to
use \sys for training an ML model.}


The core
of \sys is a \textit{training
algorithm}, implemented by 
developers using the communication 
primitives and abstractions
provided by \sys. 
An algorithm takes as input
a neural network, provided
by the end-user, and equips it 
with an algorithm-specific 
\textit{communication function}.
Specifically, the developer of
an algorithm achieves this 
by registering
this communication function 
as \textit{hooks} at different
stages of execution. One
example is to register one 
hook after the backward 
computation of each layer.
The communication function 
contains the core logic of 
a training algorithm,
which has the following signature:
\[
f((x_1, g_1)...(x_n, g_n)) \mapsto (x_1', g_1')...(x_n', g_n')
\]
where 
$(x_i, g_i)$ are 
the current model ($x_i$)
and gradient ($g_i$)
on the $i^{th}$ machine
and $(x_i', g_i')$ are
the updated model 
and gradient 
on the $i^{th}$ machine.
To implement a communication
function, the developer of 
an algorithm assumes an
MPI-like execution model.
The key difference is that 
the developer is equipped with 
not only the standard 
communication primitives
in MPI (e.g., \texttt{Allreduce}),
but also a set of 
communication primitives
provided by \sys. These 
primitives 
support system relaxations
such as compressed communications
with error compensation,
or decentralized communications.
 
When implementing the communication 
function in \sys, the developer 
provides a \textit{batched version}
of such a function, taking as input
a \textit{set} of layers. This allows
\sys to later batch the communications
automatically and optimize for its 
overlapping with the computations.
When \sys invokes this function,
it will rearrange  
parameters of all layers
into consecutive memory space 
and also pass in a \textit{flattened}
version of these layers, treat them 
as a single variable. An algorithm 
developer can decide whether her 
algorithm can use
this flattened version to avoid 
conducting communication for every layer
by communicating once for all 
the layers.
\blue{In Appendix~\ref{sec:program_algo}, we offer an example that
implements a training algorithm in \sys.}

\blue{During the runtime,} each invocation to 
the communication function
(which is triggered by the 
registered hooks)
is registered with \sys, which 
equips \sys a global view
of the workload to enable 
automatic scheduling and batching.
The key technical contribution of \sys 
is to automatically apply a series of 
optimizations for computations 
and communications. To make this happen,
the core of \sys is 
the \textit{execution optimizer},
which runs in two phases.

{\em 1. \underline{Profiling Phase.}} During the 
first forward/backward pass of the \blue{gradient descent computation},
\sys keeps a log of all 
invocations of communication functions,
executes them without any optimizations.
It then automatically: \textbf{(1. Bucketing)} groups layers into 
different buckets, whose communication 
will happen all at once; \textbf{(2. Flattening)} rearranges all 
the models and gradients of all 
layers in the same group into 
consecutive memory spaces to achieve better locality;
\textbf{(3. Scheduling)} schedules
when to conduct the communication 
of each bucket, overlapping with 
computations.

{\em 2. \underline{Execution Phase.}} For the
rest forward/backward passes of the gradient decent computation,
\sys will conduct execution over 
an automatically optimized version of
the model. By default, 
\sys conducts one communication per bucket.

\subsection{Communication Primitives}
\label{sec:sys:comm_primitive}

One key component of \sys is a collection of
communication primitives. 
All these operators follow an execution 
model similar to MPI, which take as input
$n$ tensors $x_1...x_n$ (which can store parameter, gradient, etc.), each at
a different worker, and outputs
new data products $x_1'...x_n'$, 
each at a different worker:
{
\[
op(x_1...x_n) \mapsto x_1'...x_n'
\]}

\paragraph*{\underline{Centralized, Full Precision}} 
\sys provides a simple primitive, 
$\texttt{C\_FP\_S}$, which provides the same 
functionality as the standard \texttt{Allreduce}
operator. Specifically:
{
\[
\texttt{C\_FP\_S}(x_1...x_n)  \mapsto x_1'...x_n' ~~~\implies~~~ \forall i \in [n].~~~x_i' = \sum_j x_j
\]}%
We use this notation to express that,
the effect of the $\texttt{C\_FP\_S}$ operator 
is to calculate the sum of all local 
replicas, $\sum_j x_j$, and make it accessible to 
all workers.

\paragraph*{\underline{Centralized, Low Precision}}
Communication compression has attracted intensive 
interests recently, given that many deep neural networks
are tolerant to aggressive 
lossy compression of its gradient~\cite{alistarh2016qsgd,zhang2017zipml,bernstein2018signsgd,wen2017terngrad,wangni2018gradient,alistarh2018convergence,wang2018atomo,wang2017efficient,ivkin2019communication,tang2019doublesqueeze}.
\sys provides the \texttt{C\_LP\_S}
primitives for this purpose. Specifically:
{
\begin{align*}
~ & \texttt{C\_LP\_S}(x_1...x_n,\delta_1...\delta_n,\epsilon_1...\epsilon_n)  \mapsto  x_1'...x_n',\delta_1'...\delta_n',\epsilon_1'...\epsilon_n' \\
\implies & \forall i \in [n]. x_i' = Q\left(\sum_j Q(x_j - \delta_j) - \epsilon_i\right) \\
~ & \forall i \in [n]. \delta_i' = x_j - \delta_j - Q(x_j - \delta_j) \\
~ & \forall i \in [n]. \epsilon_i' = 
\sum_j Q(x_j - \delta_j) - \epsilon_i -
Q \left(\sum_j Q(x_j - \delta_j) - \epsilon_i\right)
\end{align*}}%
where $Q$ is the lossy compression function, specified by 
the developer and \texttt{C\_LP\_S}
supports a general form of communication 
compression with error compensation~\cite{tang2019deepsqueeze,tang2019doublesqueeze}. Note that setting $\delta_i$ and $\epsilon_i$ to \texttt{None} will
disable error compensation and gives
{
\begin{align*}
~ & \texttt{C\_LP\_S}(x_1...x_n,\texttt{None},\texttt{None})  \mapsto  x_1'...x_n' \implies \forall i \in [n].~~~x_i' = Q\left(\sum_j Q(x_j) \right) 
\end{align*}}%
Intuitively, $\delta_i$ and $\epsilon_i$ keep the
error caused by last iterations' compression.  
The convergence efficiency introduced by error compensated methods is quite robust to the compression.
This technique is especially helpful when
the compression function is relatively aggressive (e.g., top-K compression~\cite{stich2018sparsified,alistarh2018convergence}).


\paragraph*{\underline{Decentralized, Full Precision}} \sys also supports decentralized communication, which gets rid of the latency bottleneck for model synchronization --- instead of synchronizing among all $n$ workers in the cluster, each worker only sends the update to its neighbors. For example, according to a ring-based topology, the neighbors of a worker include its immediate left and immediate right workers in the ring. Formally, \sys's decentralized full precision communication primitive \texttt{D\_FP\_S} can be formalized as below:
{
\[
\texttt{D\_FP\_S}(x_1...x_n)  \mapsto x_1'...x_n' ~~~\implies~~~ \forall i \in [n].~~~x_i' = \sum_{j \in \mathcal{N}(i)} x_j
\]}%
where $\mathcal{N}(i)$ is the set of workers that
are neighbors of worker $i$. Note that  $\mathcal{N}(i)$ 
is an input to \texttt{D\_FP\_S}, which can be a deterministic 
function (e.g., fixed ring topology) or a 
randomized function.

\paragraph*{\underline{Decentralized, Low Precision}} \sys also provides the primitive \texttt{D\_LP\_S} for decentralized low precision communication:
{
\[
\texttt{D\_LP\_S}(x_1...x_n)  \mapsto x_1'...x_n' ~~~\implies~~~ \forall i \in [n].~~~x_i' = \sum_{j \in \mathcal{N}(i)} Q(x_j)
\]}

\paragraph*{\underline{Discussion: Supporting Asynchronous Algorithms}}
\blue{Regarding the asynchronous algorithms,}
the current version of \sys does not provide 
any asynchronous version of these primitives,
instead, it supports asynchronous algorithms
using these synchronous primitives 
as follows. An algorithm can implement two 
concurrent threads, one 
deals with computation and another deals with 
communications. These two threads do not wait
for each other. This provides an implementation 
of many asynchronous algorithms~\cite{zheng2017asynchronous, lian2018asynchronous,de2017understanding},
summarized in Table~\ref{fig:landscape}. 
It can also enable
implementations for LocalSGD~\cite{lin2019don}
and model averaging~\cite{yu2019parallel}.
It is interesting to further explore the benefits
of providing asynchronous version of 
primitives, which we leave as future work.

\subsection{\blue{Comparisons with PS and Allreduce}}
\label{sec:discussion}

\blue{
We see \sys as a \textit{natural extension} of two popular 
existing paradigms, i.e., parameter server and 
\texttt{Allreduce}.
The design of \sys's communication primitives is inspired 
by challenges that we faced when trying to implement
all eight different communication patterns in Table~\ref{fig:landscape}
when directly using these paradigms, discussed as follows.
}

\paragraph*{\blue{Parameter Server}} 
\blue{
Considering the centralized, low precision communication pattern
(\texttt{C\_LP\_S}), we found it could be unnatural to implement 
such a pattern by directly using the \texttt{put/get} abstraction provided by a parameter server. The fundamental challenge is that the error compensation step 
requires us to keep a state (to remember the error) on the 
server side and continue to conduct accumulation and quantization
for each \texttt{get} request. The \texttt{C\_LP\_S} primitive 
is an extension of the parameter server abstraction 
for this purpose. A more fundamental problem is 
to support decentralized algorithms (\texttt{D\_FP\_S}) using 
the parameter server abstraction --- we have to explicitly 
specify the communication topology. The \texttt{D\_FP\_S}
primitive is an extension for this purpose.
}

\paragraph*{\blue{Allreduce}} 
\blue{Similarly, using \texttt{AllReduce} operators
in MPI also faces challenges when implementing 
a centralized, low precision communication pattern.
It is not clear how to keep track of the quantization 
error made by \texttt{Allreduce} at each communication step 
and compensate it in the next iteration. Similarly,
it can also be unnatural to support the decentralized communication pattern
using existing operators in MPI.}

\subsection{Implementations of Primitives}

\blue{For the centralized primitives, \sys adopts a ScatterReduce communication pattern~\cite{zhang2019mllib}.} 
Specially, the target tensor is divided into $n$ partitions, where $n$ is the number of workers.
The $i$-th worker is responsible for aggregating the $i$-th partition.
Since the underlying communication library NCCL does not
provide a ScatterReduce primitive,
we implement this primitive using the basic \textit{send} and \textit{recv} NCCL operators.
Each worker 1) partitions local tensor,
2) sends partitions to corresponding workers, 
3) receives responsible partitions from other workers,
4) merges received partitions, and
5) sends merged partition to other workers.
ScatterReduce communication pattern can take advantage of the aggregated bandwidth of all workers (like Allreduce), and support compression techniques (unlike Allreduce).
The low precision primitive $\texttt{C\_LP\_S}$ leverages the ScatterReduce communication to incorporate two phases of compression. 
Note that, the compression and decompression procedures can be
combined with error compensation technique to reduce 
information loss (see semantics in Section~\ref{sec:sys:comm_primitive}).

\blue{Unlike centralized training in which
all workers are involved in the communication, workers in decentralized training only
communicates with one or a few peers.} \sys engineers two mechanisms to allocate peers --- \textit{ring} and \textit{random}.
The \textit{ring} strategy gives successive ranks to workers and organizes all workers as a ring.
The rank-$i$ worker only communicates with two neighboring peers --- rank-$(i-1)$ and rank-$(i+1)$.
Alternatively, the \textit{random} strategy randomly chooses a peer for each worker.
The low precision primitive $\texttt{D\_LP\_S}$ uses the same peer selection and communication procedure as $\texttt{D\_FP\_S}$.
The difference is that $\texttt{D\_LP\_S}$ uses the compression function $Q$ to compress the tensor before sending and  decompress tensor after receiving.

\subsection{\sys Optimization Framework}
\label{sys_optimization}

The central component of \sys is its 
\textit{execution optimizer}. Given 
a neural network as input, a 
training algorithm (e.g., QSGD)
would utilize a sequence of 
communication primitives 
during the computations of each layers.
The goal of \sys's execution optimizer
is to automatically schedule and 
optimize these computations and 
communications.
We explore the following techniques in \sys . 

\paragraph*{\underline{Overlap Communication and Computation}}
\blue{Overlapping communication and computation is one central optimization to speedup distributed DP-SG.} Not limited to the DP-SG algorithm, \sys is able to overlap communication primitives along with the computation
for other algorithms.
\sys automatically analyzes the computation graph that includes the in-place tensor operations and tensor communication primitives. 
Compared with existing systems, \sys considers more sophisticated scheduling. In vanilla DP-SG, the optimization can only hide the Allreduce communications inside the gradient computation,
\sys is responsible for scheduling additional elements, such as compression/decompression and the model update computations specified by the optimization algorithms (E.g., Figure~\ref{fig:relaxation_patterns}).

\paragraph*{\underline{Tensor Bucketing and Memory Flattening}} 
\blue{In order to boost the efficiency of communication and parallel computation, 
fusing small tensors into buckets is an essential step} --- frequently calling the communication paradigms to transfer small fragments of parameter is far from ideal \blue{in terms of} fully utilizing the network bandwidth.
As so, the bucketing trick is adopted in both Horovod and PyTorch-DDP. 
However, their bucketing schema simply considers the Allreduce operation as the cost in the heuristic. By contrast, since \sys supports much more communication patterns, the bucketing strategy needs to consider beyond the Allreduce. Once we split the computation graph into buckets, \sys conducts fusion over the buckets. 
After determining the partition of buckets in the first run of backward propagation, \sys would carefully align parameters (e.g., model parameters, gradients) within a bucket into a continuous memory space. Then this flatten view of the parameters is leveraged for all the executions. For example, the low-precision compression/decompression lambda is directly applied over a flatten view of the bucket instead of individual parameters; the SG based optimizer for model update \blue{is also conducted at the level of buckets} (Apex~\cite{apex} from NVIDIA also uses a similar optimization). Note that this flatten view can utilize the parallelism offered by the computation unit more effectively.   


\paragraph*{\underline{Hierarchical Communications}}
\blue{Last but not least, the communication of \sys can be conducted hierarchically.} This is particularly useful when dealing with the heterogeneity in network connections, e.g., the bandwidth between GPUs within a server is much higher than the bandwidth between servers. 
Therefore, \sys communicates hierarchically in two levels: intra-node and inter-node, and optimize the implementation of communication primitives based on this abstraction. 
For example, the centralized low-precision primitive (\texttt{C\_LP\_S}) can be optimized as first aggregating tensors over the local workers inside each node \textit{without} compression, then performing inter-node aggregation over the leader workers \textit{with} compression, and finally letting each leader worker broadcast aggregated data within the node.
Notice that this optimization can potentially change the semantics of the communication primitives. 

\section{Evaluation}
\label{sec:eval}

We conduct extensive experimental study around three 
hypotheses:
\begin{itemize}
\item \textit{\sys is able to provide significant performance improvements over state-of-the-art systems in terms of end-to-end training time and scalability, over realistic 
industrial-scale infrastructure.}
\item \textit{Different algorithms that \sys support 
provide benefits for different models and datasets under different network conditions. It is thus important for \sys to support all 
these algorithms.}
\item \textit{\sys's automatic execution optimizer effectively optimizes the execution of various distributed training algorithms.}
\end{itemize}

\subsection{Experimental Setting}

\paragraph*{\underline{Infrastructure}} 
All experiments are conducted 
on 16-GPU instances, each of which is equipped with 8 NVIDIA V100 32GB GPUs interconnected by NVLink. We consider three different network conditions
following how V100 GPU machines (\texttt{p3.8xlarge}, \texttt{p3.16xlarge}, \texttt{p3dn.24xlarge}) are connected on AWS:
10Gbps, 25Gbps, and 100Gbps, with TCP/IP connections. The default bandwidth we are using is 100Gbps without specification.

\paragraph*{\underline{Competing Systems}}
We compare the performance of \sys with three state-of-the-art systems. \textbf{PyTorch-DDP}~\cite{li13pytorch},  Pytorch's default solution of distributed data parallelism learning. 
\textbf{Horovod}~\cite{sergeev2018horovod}, a distributed learning framework developed by Uber. 
\textbf{BytePS}~\cite{jiang2020unified}, 
a distributed learning platform developed by
ByteDance. Both PyTorch-DDP and Horovod relies on 
MPI Allreduce for communication while BytePS
uses parameter servers.
Horovod and PyTorch-DDP also supports \textbf{fp16 
gradient compression} via the fp16 support in NVIDIA NCCL, 
which we also compare with.

\begin{sloppypar}
\paragraph*{\underline{Datasets and Tasks}}
We use five learning tasks, covering different modalities and both standard benchmarks and 
production datasets at Kwai Inc:
\textbf{(1)} \textbf{Image}: (ImageNet~\cite{imagenet_cvpr09}, VGG16~\cite{simonyan2014very});
\textbf{(2)} \textbf{Text}: (SQuAD~\cite{rajpurkar2016squad}, BERT-LARGE finetune~\cite{devlin2018bert});
\textbf{(3)} \textbf{Text}: (Kwai Dataset, BERT-BASE finetune~\cite{devlin2018bert});
\textbf{(4)} \textbf{Speech}: (AISHELL-2 \cite{du2018aishell}, Transformer);
\textbf{(5)} \textbf{Image+Text}: (Kwai Dataset, LSTM~\cite{hochreiter1997long}+AlexNet~\cite{krizhevsky2012imagenet}).
Table~\ref{tb:models} summarizes the model size and FLOPs.
\end{sloppypar}

\begin{table}[h]
\centering
\begin{tabular}{c c c c c c}
\hline
 & VGG16 & BERT-LARGE  & BERT-BASE & Transformer & LSTM+AlexNet \\
\hline
\# Parameters & 138.3M & 302.2M & 85.6M & 66.5M & 126.8M\\

\# FLOPs  &31G & 232G & 22G & 145G & 97.12G \\
\hline
\end{tabular}
\caption{Model Characteristics}
\label{tb:models}
\end{table}

\paragraph*{\underline{\sys Algorithms}}
We implemented six algorithms in \sys.
\textbf{Allreduce}, the standard DP-SG algorithm, implemented with \texttt{C\_FP\_S} primitive.
\textbf{QSGD}~\cite{alistarh2016qsgd}, a quantized (8-bit) DP-SG algorithm, implemented with \texttt{C\_LP\_S} primitive without error compensation.
\textbf{1-bit Adam}~\cite{tang20211}, a quantized (1-bit) distributed learning algorithm, implemented with by \texttt{C\_LP\_S} primitive with error compensation.
\textbf{Decen-32bits}, a decentralized training algorithm with the random probing method to exchange the model parameters in each iteration, implemented with \texttt{D\_FP\_S}.
\textbf{Decen-8bits}~\cite{tang2018communication}, a ring-based decentralized training algorithm with quantization, implemented with \texttt{D\_LP\_S}.
\textbf{Async}, asynchronous centralized DP-SG.

\begin{figure*}[t!]
	\subfigure[VGG16]{
		\scalebox{0.3}{
			\includegraphics[width=1\linewidth]{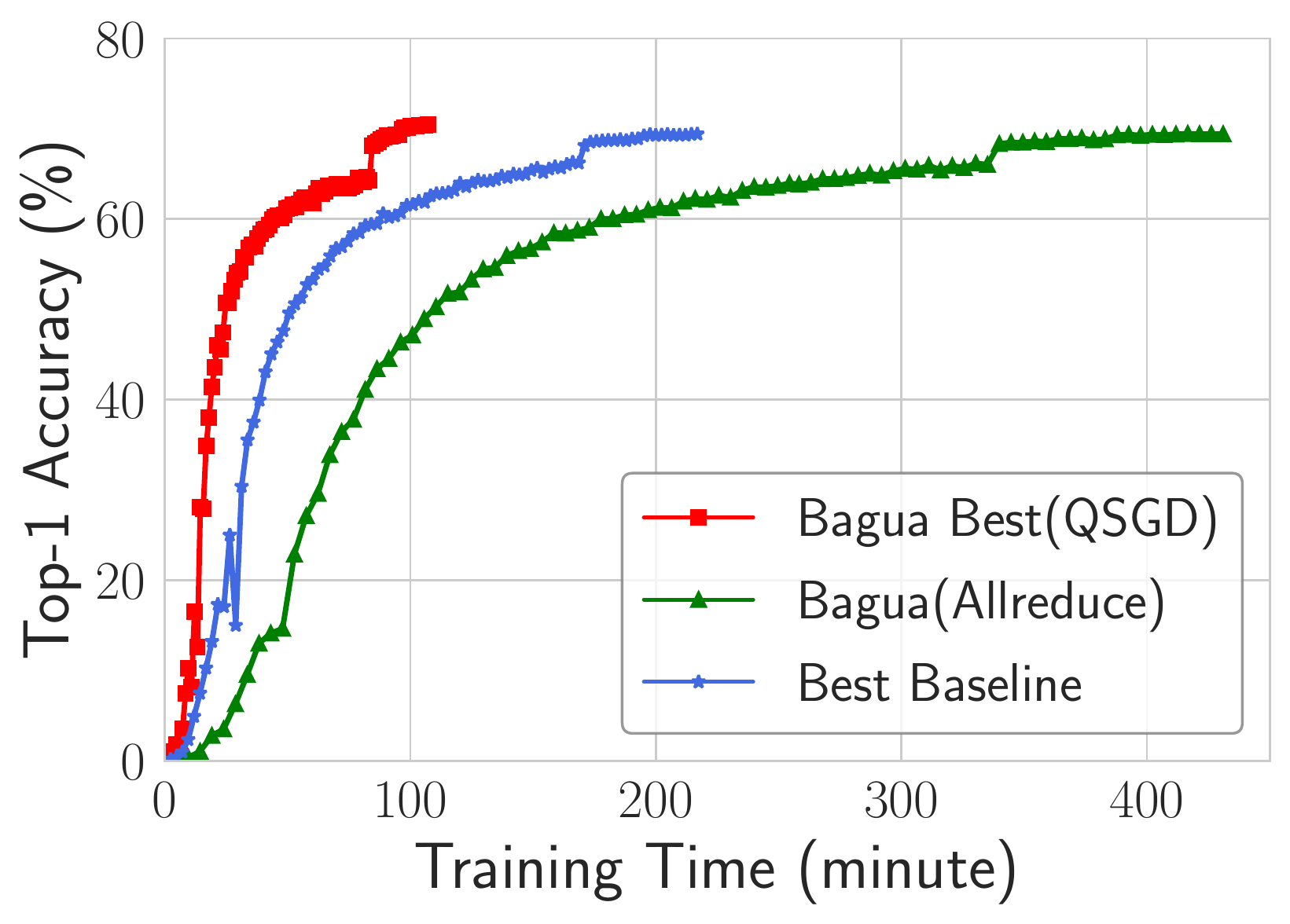}
	}}
	\subfigure[BERT-LARGE Finetune]{
		\scalebox{0.3}{
			\includegraphics[width=1\linewidth]{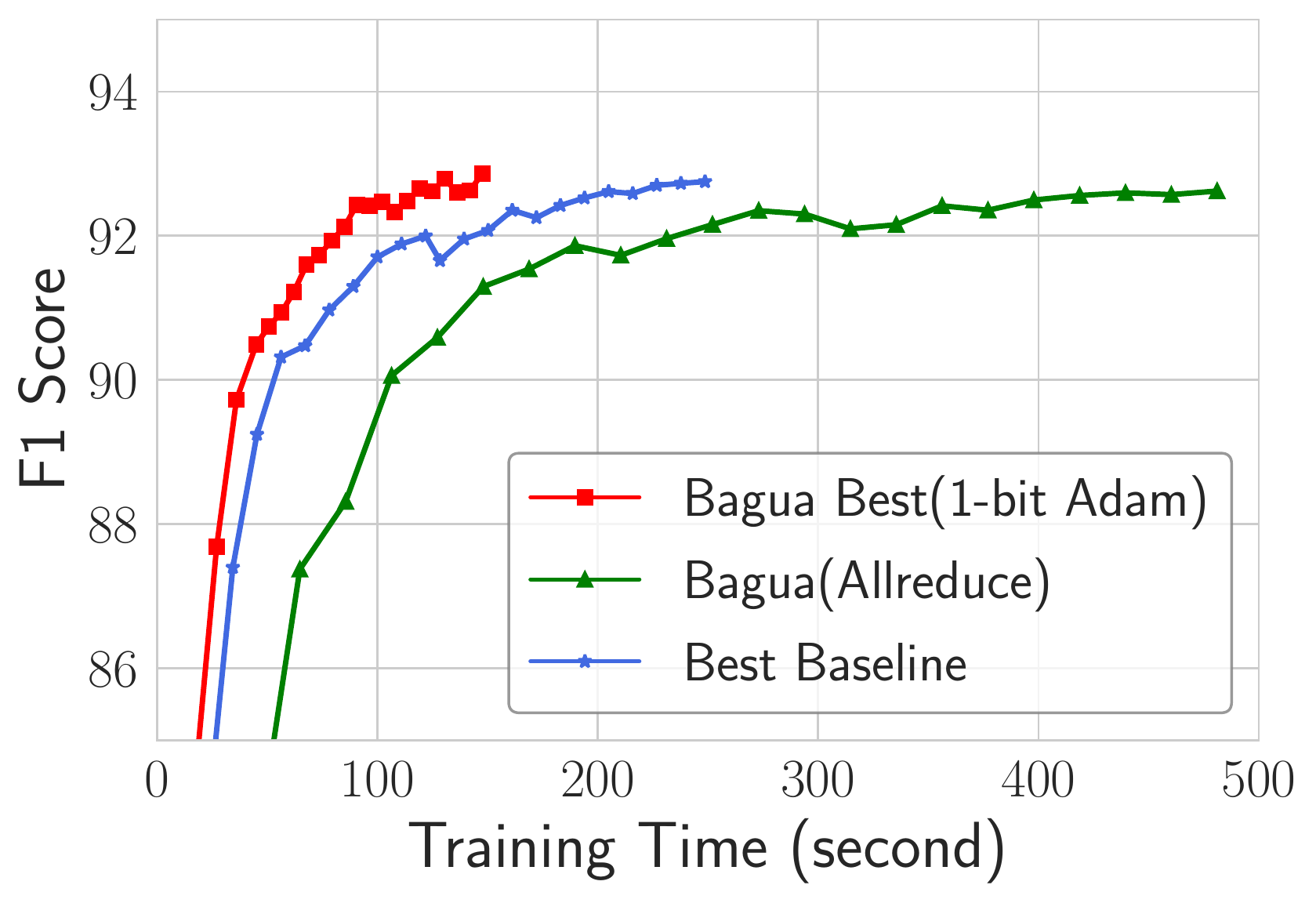}
	}}
	\subfigure[BERT-BASE Finetune]{
		\scalebox{0.3}{
			\includegraphics[width=1\linewidth]{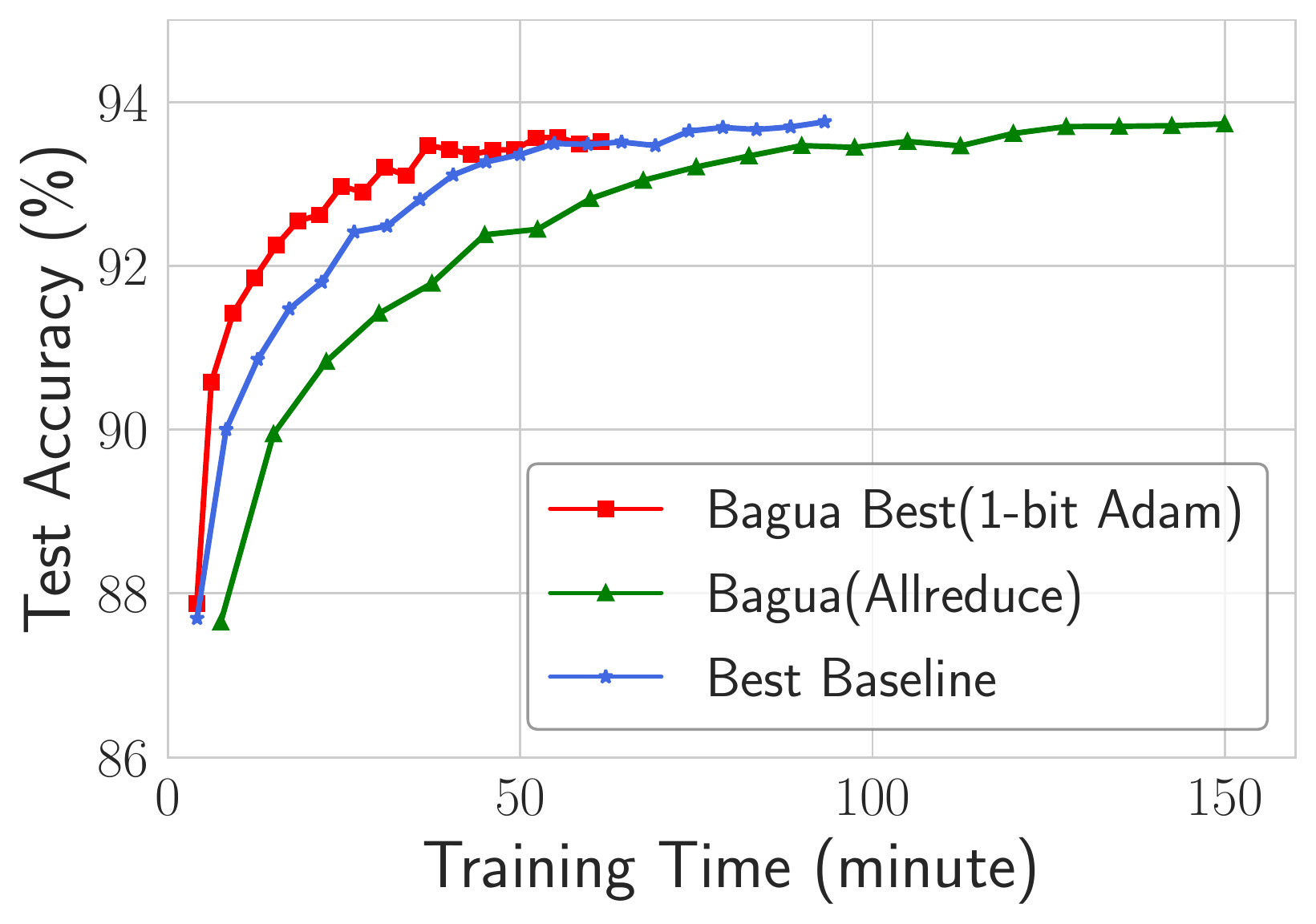}
	}}
    \subfigure[Transformer]{
		\scalebox{0.3}{
			\includegraphics[width=1\linewidth]{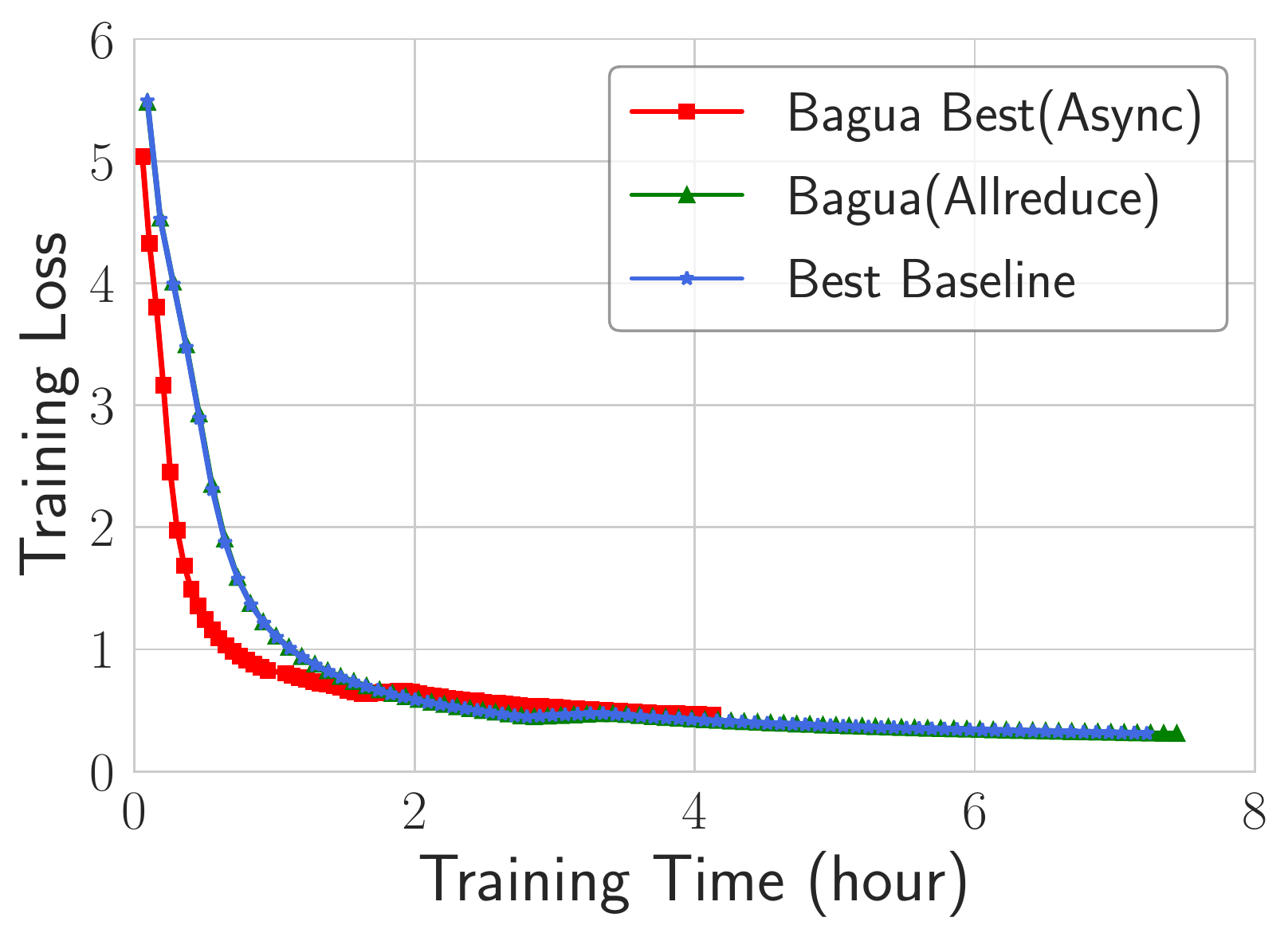}
	}}
    \subfigure[LSTM+AlexNet]{
		\scalebox{0.3}{
			\includegraphics[width=1\linewidth]{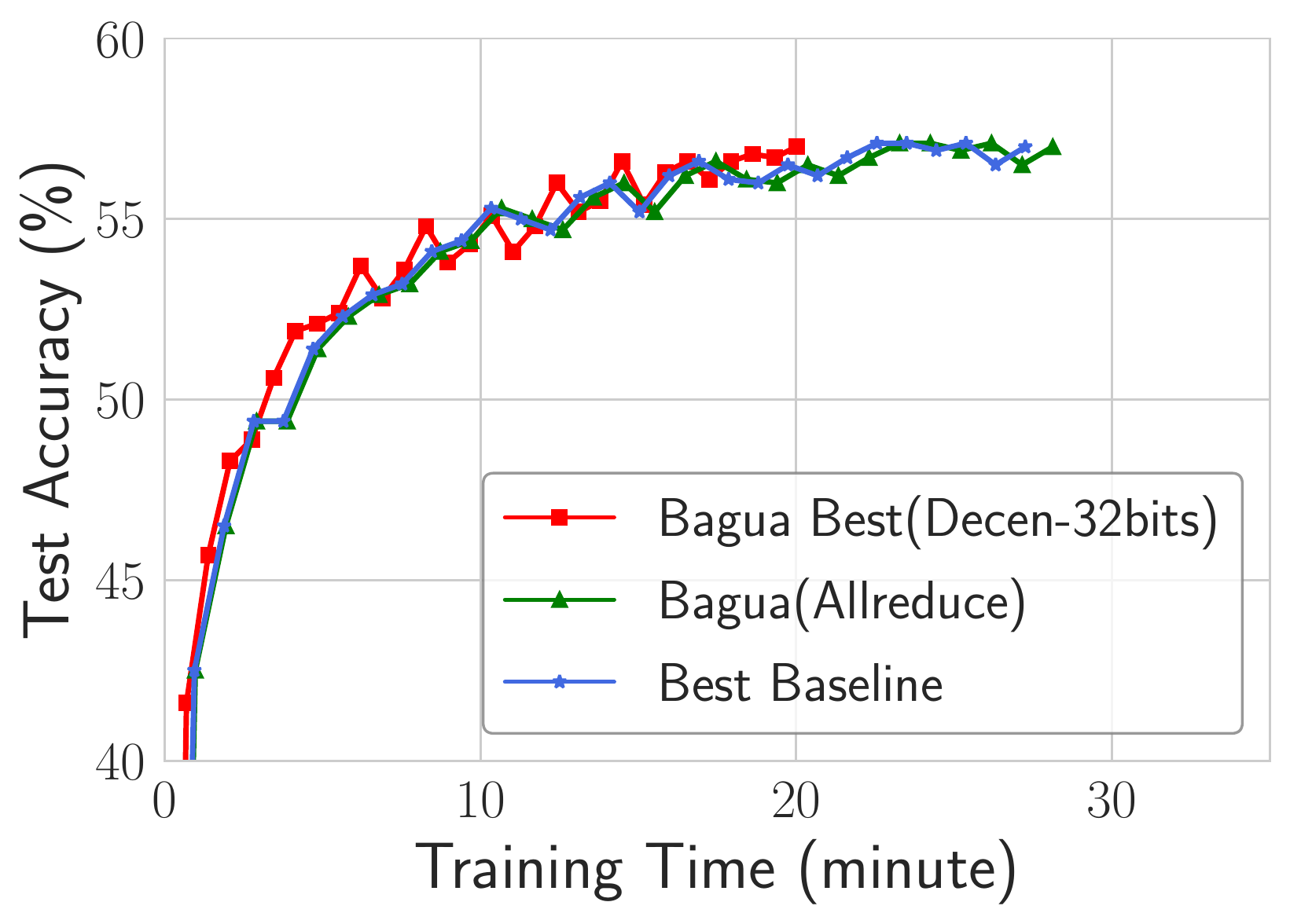}
	}}
	\caption{\blue{End-to-end performance of \sys and the best competing baseline. 
	Over all five tasks, Horovod-16bits is the \textit{best} of \{Torch-DDP, Horovod-32bits, Horovod-16bits, BytePS\}. We show the performance of \sys Allreduce and the optimal \sys algorithm selected for each task. The bandwidth of inter-machine network is 10Gbps.}}
  \label{fig:end2end}
\end{figure*}

\begin{figure*}[t!]
	\subfigure[VGG16]{
		\scalebox{0.3}{
			\includegraphics[width=1\linewidth]{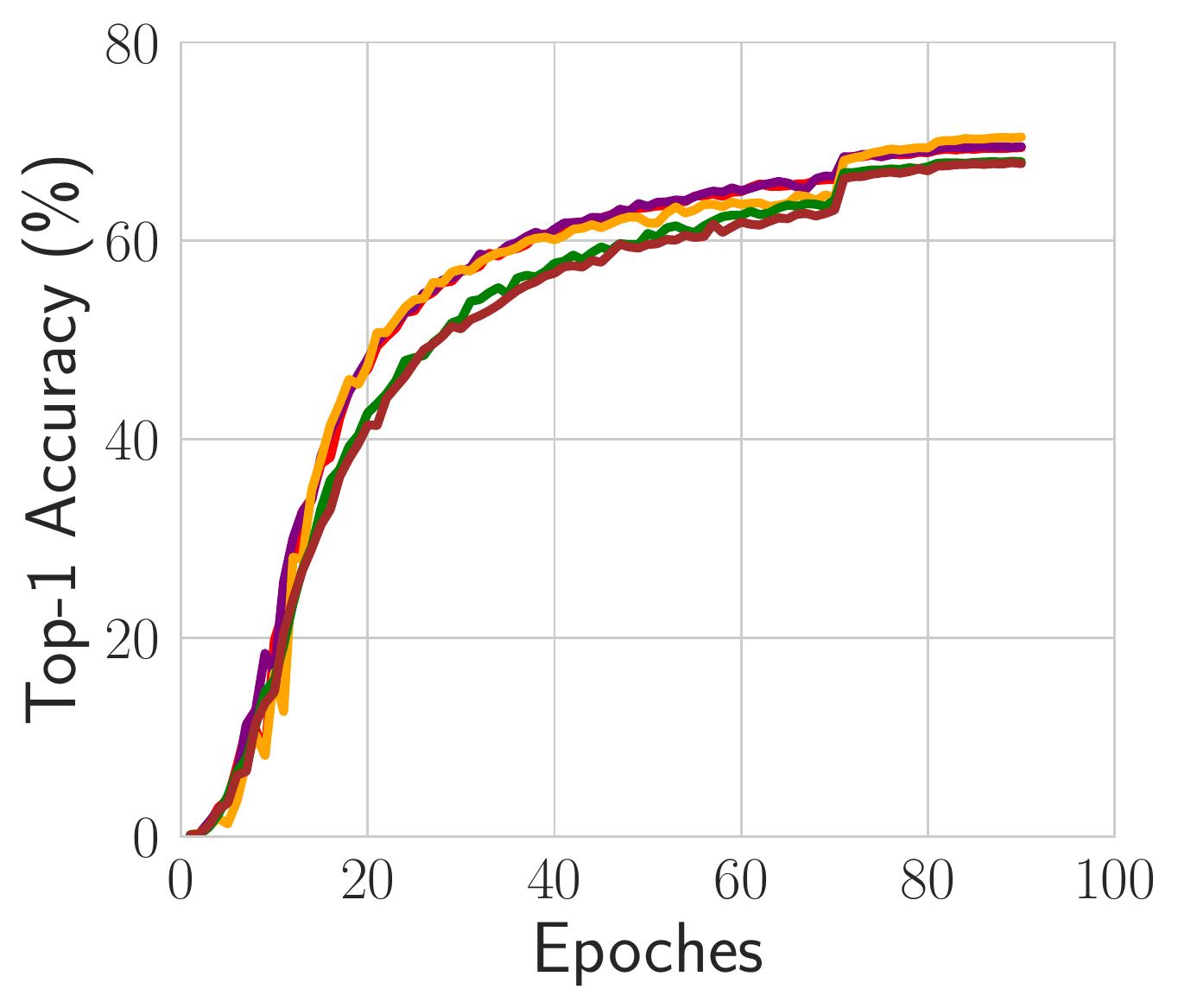}
	}}
	\subfigure[BERT-LARGE Finetune]{
		\scalebox{0.3}{
			\includegraphics[width=1\linewidth]{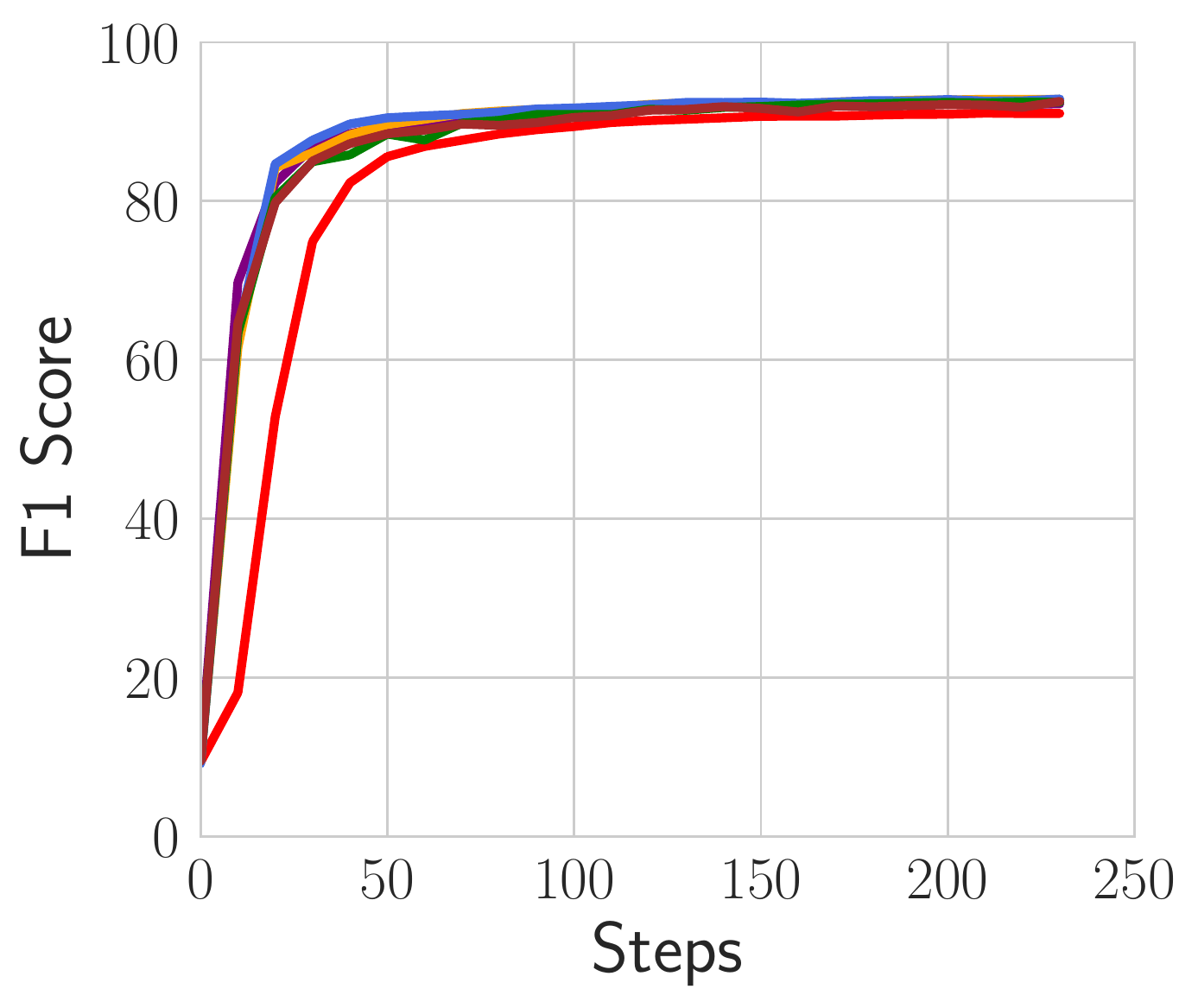}
	}}
	\subfigure[BERT-BASE Finetune]{
		\scalebox{0.3}{
			\includegraphics[width=1\linewidth]{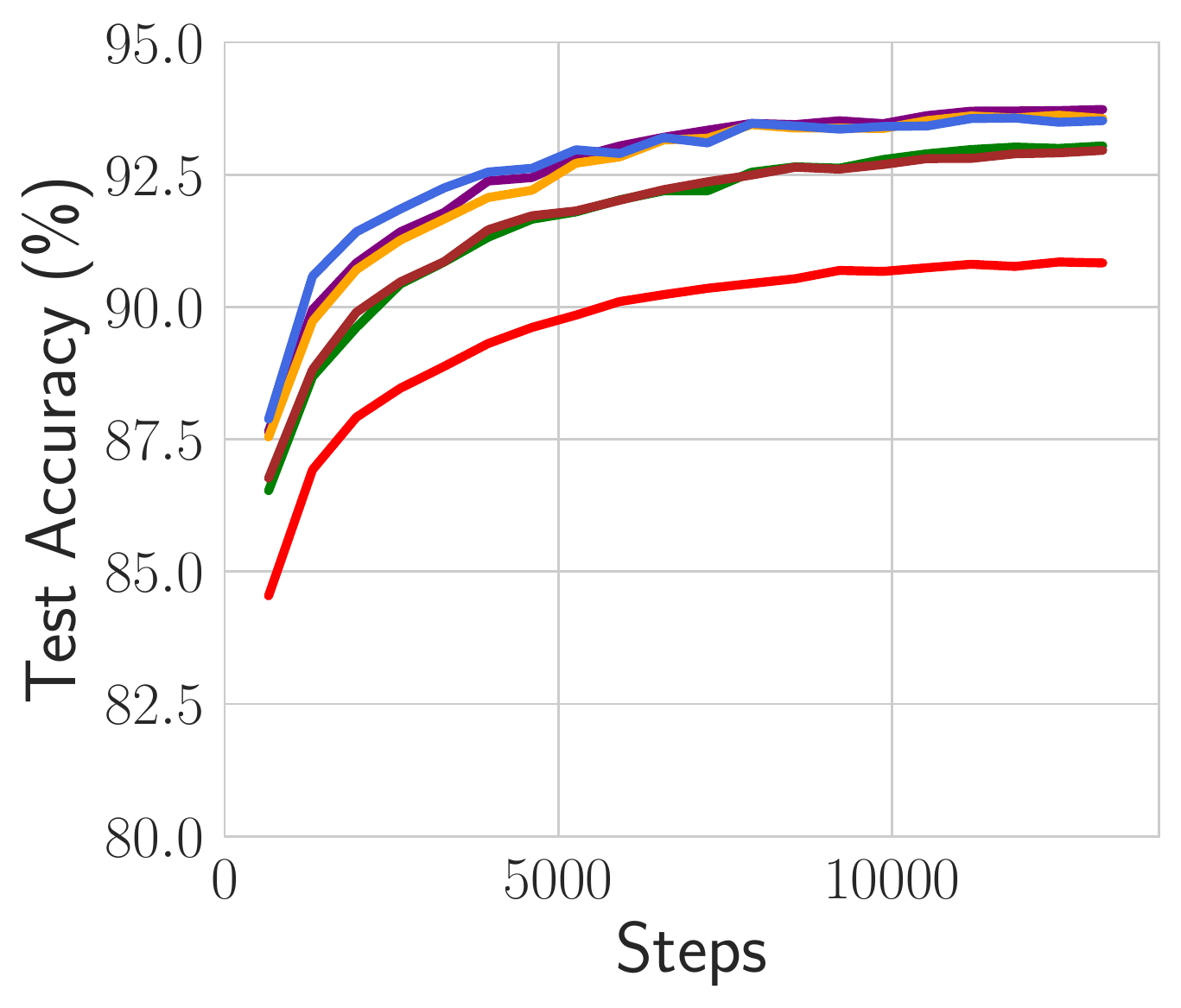}
	}}
    \subfigure[Transformer]{
		\scalebox{0.3}{
			\includegraphics[width=1\linewidth]{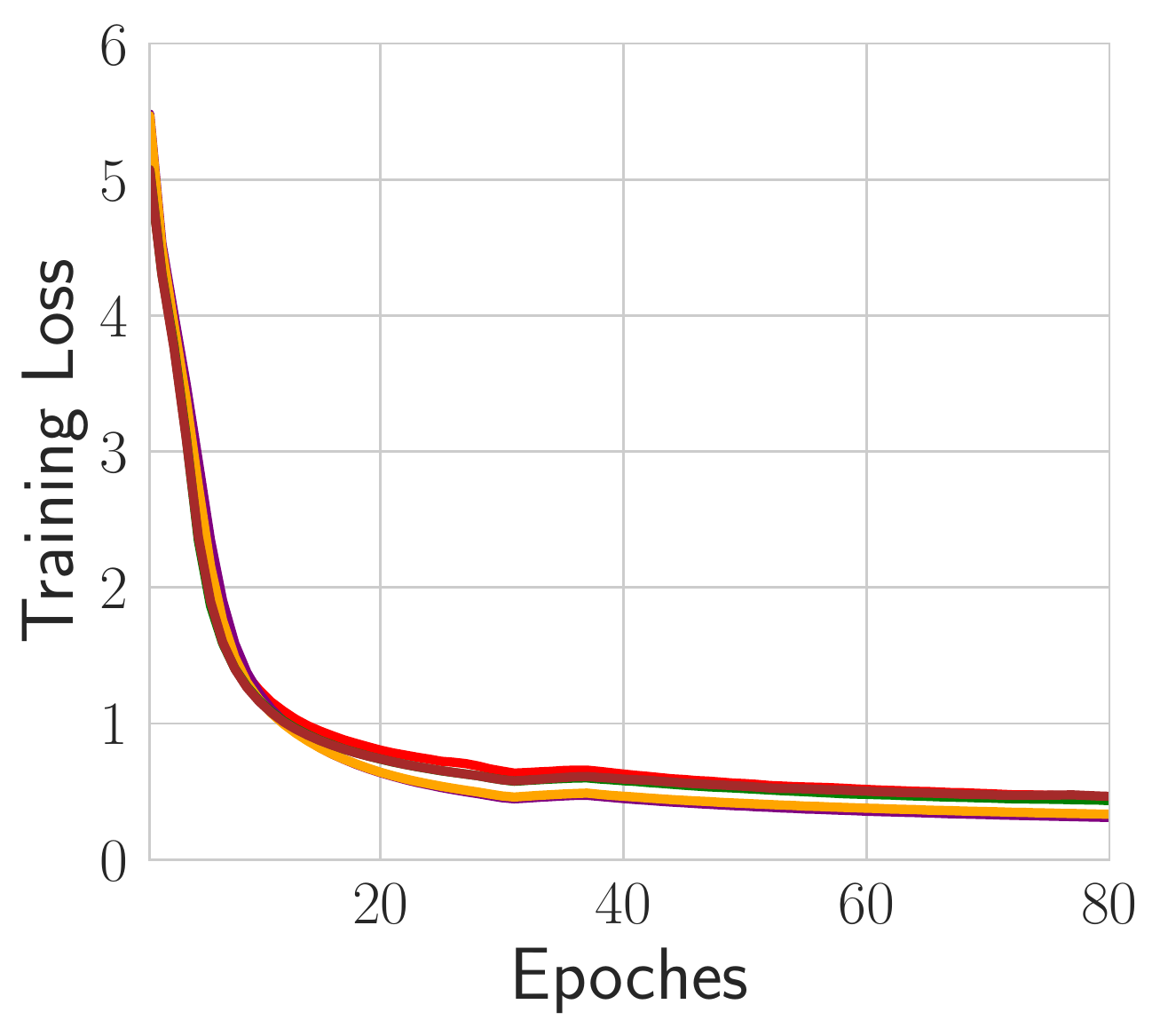}
	}}
    \subfigure[LSTM+AlexNet]{
		\scalebox{0.3}{
			\includegraphics[width=1\linewidth]{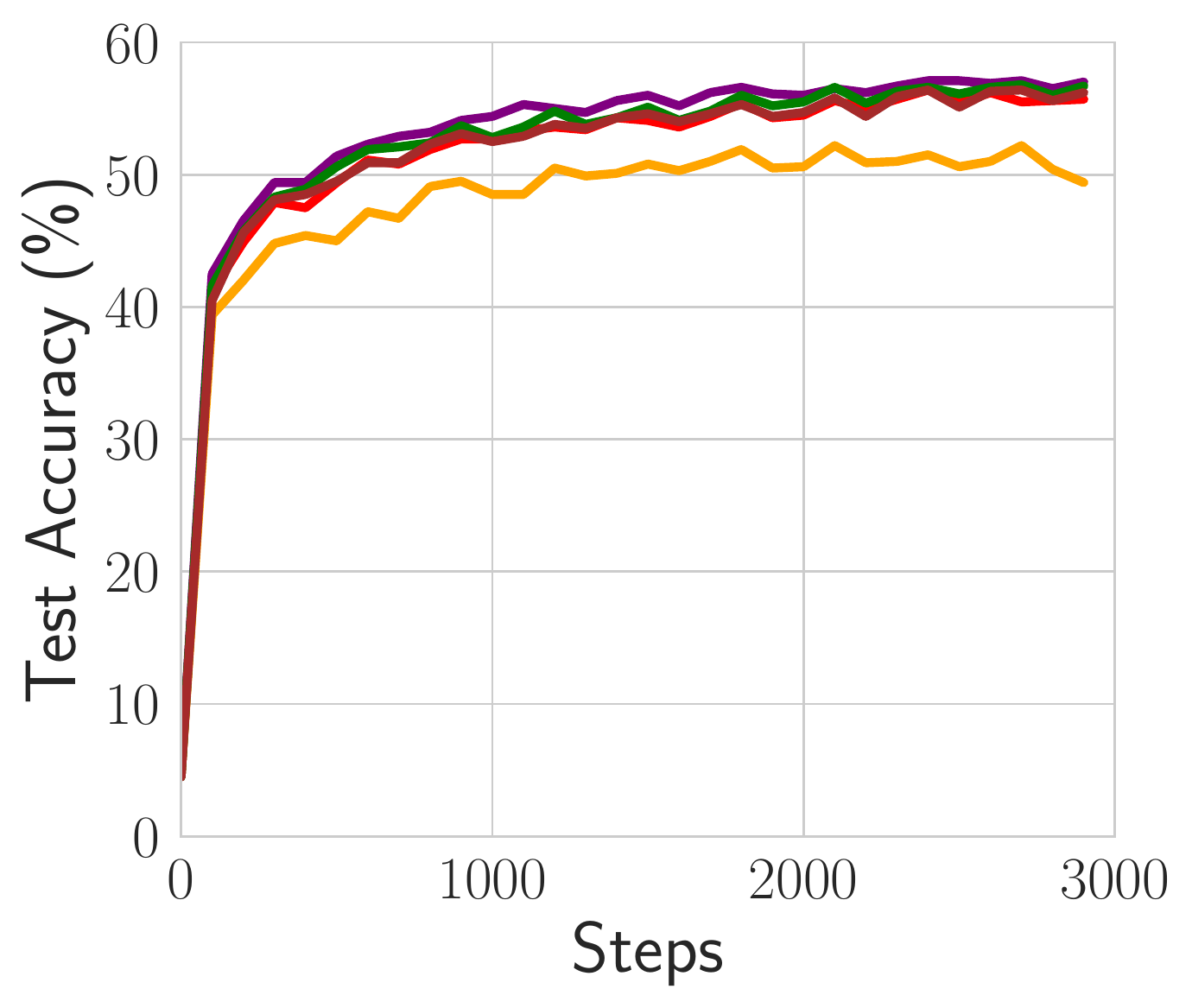}
	}}
	\subfigure{
		\scalebox{0.3}{
			\includegraphics[width=1\linewidth]{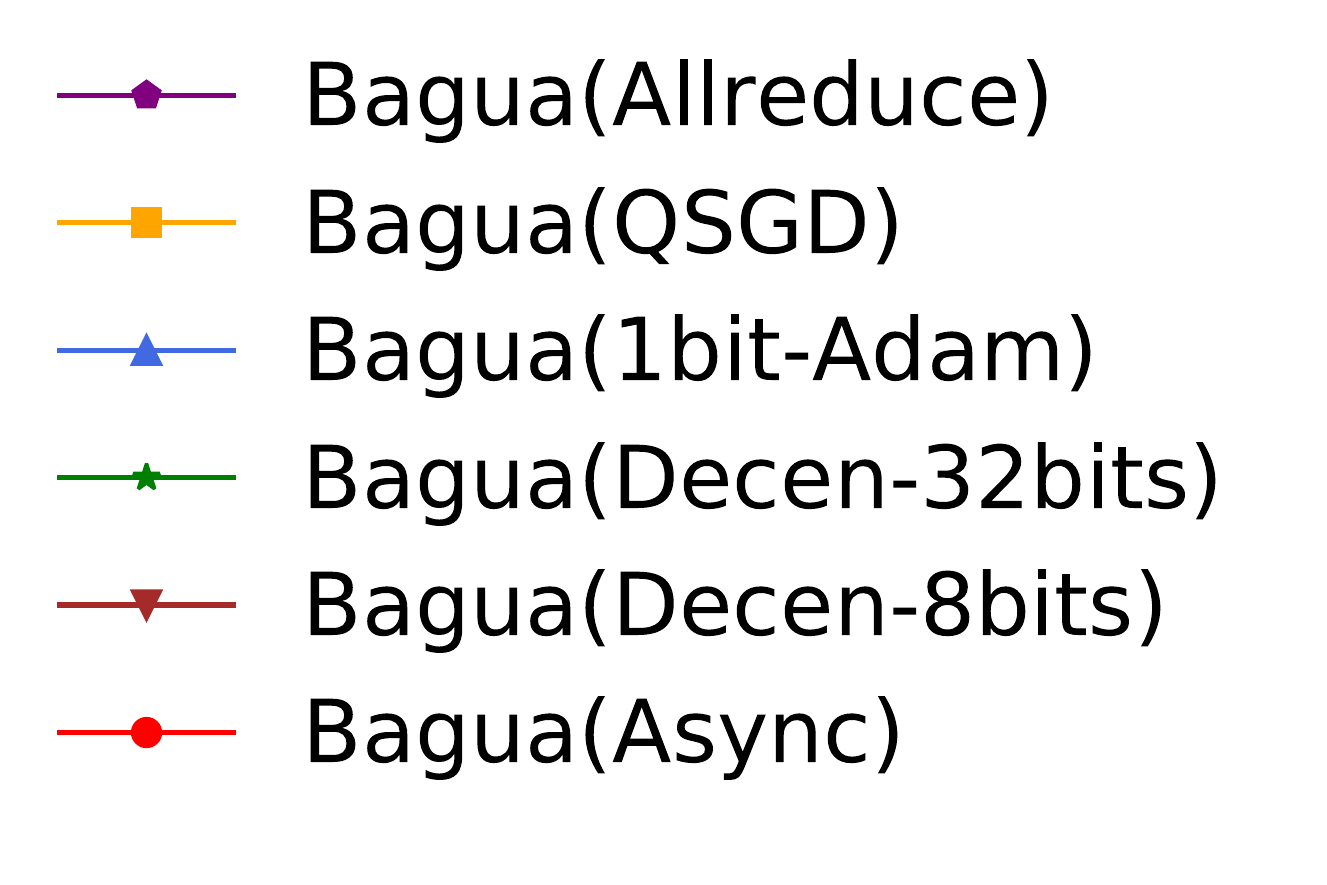}
	}}
	\caption{\blue{Convergence of different algorithms}}
  \label{fig:tradeoff-conv}
\end{figure*}

	

\subsection{End-to-end Comparison with SOTAs}
\blue{We first conduct end-to-end experiments over the network with 10Gbps bandwidth, following 
the setup of V100 instance \texttt{p3.8xlarge} on AWS.
In Figure \ref{fig:end2end}, we compare \sys with Horovod-16bits, which is the best of all the baseline systems in this setting. We also present the results over 100Gbps network in the Appendix~\ref{sec:100g}. We report the results of \sys Allreduce and the best algorithm in \sys we select for each task by considering both the accuracy and efficiency. As we can see, \sys can be up to $2\times$ faster than the best competing system, while still guaranteeing the same model accuracy. \sys Allreduce is the slowest one in this case because it has the largest amount of data transition. These results also reflect the effectiveness and necessity of supporting various algorithms.}

\blue{
There are multiple reasons behind the speedup of \sys. Firstly, \sys is able to benefit from various communication-efficient algorithms. For example, \sys (QSGD-8bits and 1-bit Adam) can compress the data communication much more aggressively than Horovod-16bits, \sys (Decentralized 32bits and 8bits) could have much less communication connections than Allreduce-based systems, and \sys (Async) can make the communication totally independent of the computation. Therefore, the overall communication overhead is highly reduced by \sys. Besides, \sys has a well-optimized execution pipeline for the computation and communication, including tensor flattening and bucketing, memory management, overlapping and so on. All the algorithms implemented in \sys are automatically benefiting from these system optimizations.}

\begin{figure}[t!]
	\subfigure[Bandwidth]{
		\scalebox{0.45}{
			\includegraphics[width=1\linewidth]{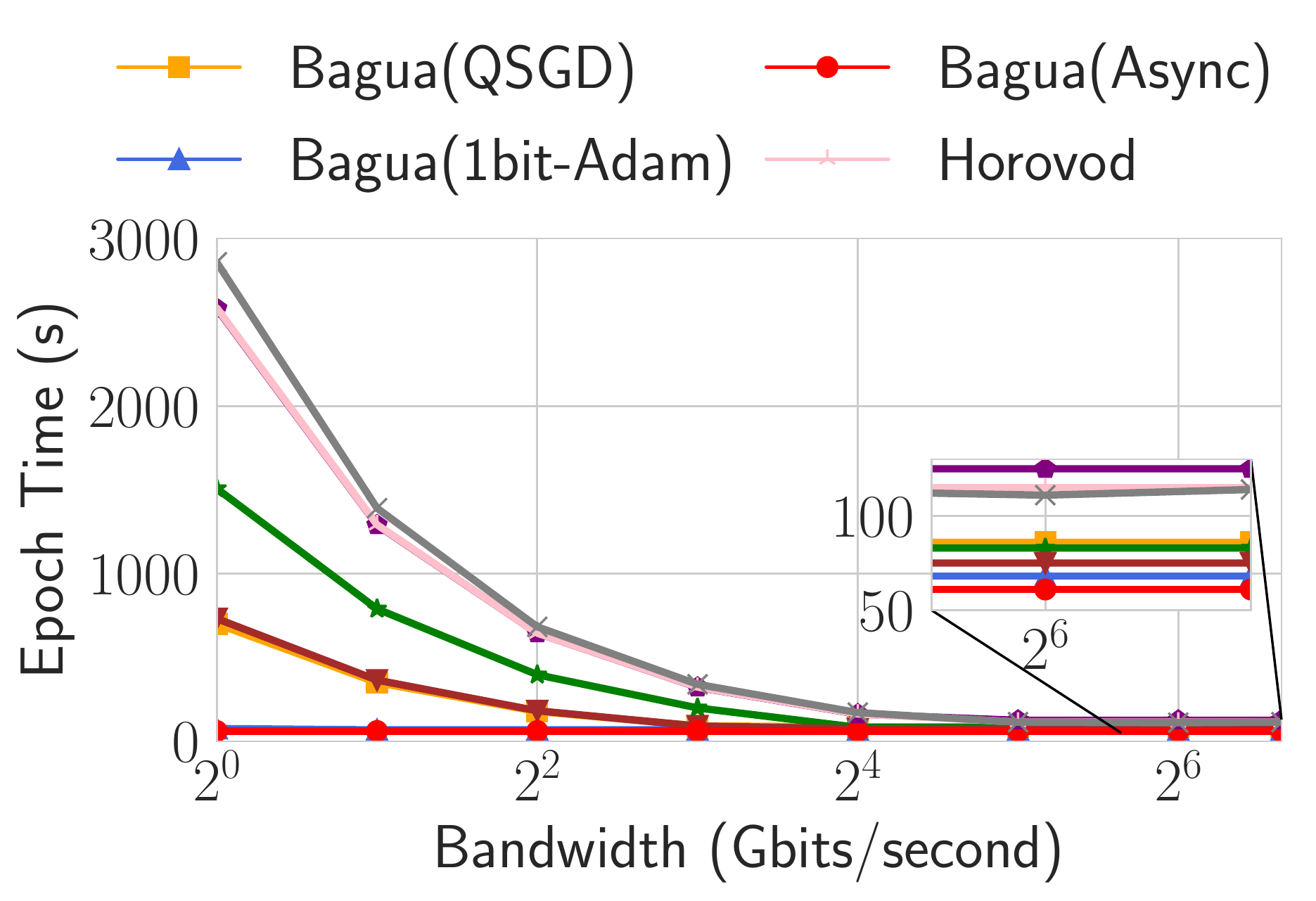}
	}}
	\subfigure[Latency]{
		\scalebox{0.45}{
			\includegraphics[width=1\linewidth]{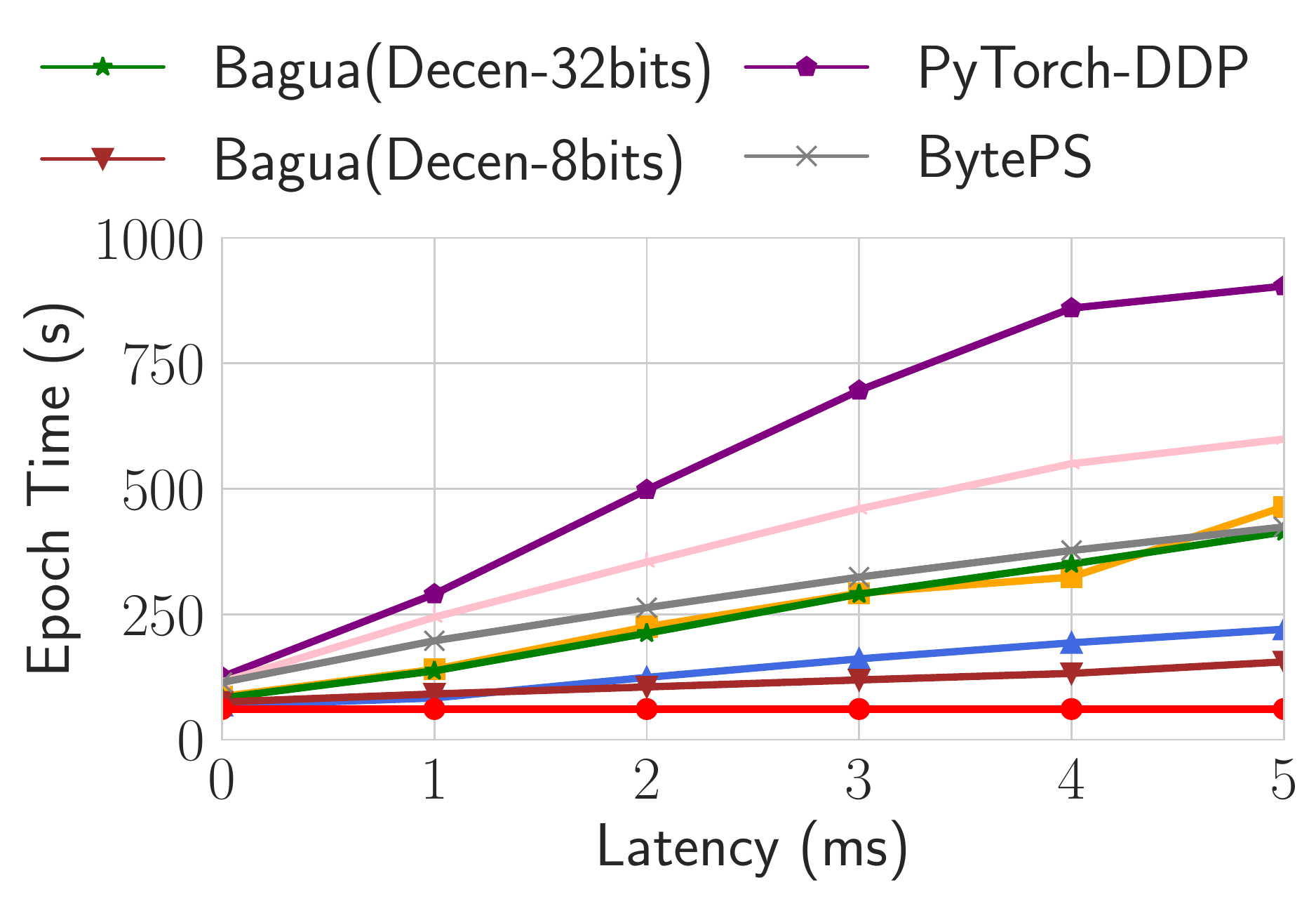}
	}}
	\caption{Epoch time under different network conditions, BERT-LARGE Finetune}
  \label{fig:tradeoff-network}
\end{figure}

\subsection{\blue{Trade-off of \sys Algorithms}}
\label{sec:tradeoff}
\blue{
By supporting a diverse collection of algorithms,
\sys provides users flexibility to accommodate 
different tasks, network conditions (in terms of latency and throughput), and worker heterogeneities. As we will see, there is no one-size-fits-all algorithm in \sys that is always optimal for all these situations. Instead, we see an interesting trade-off space of \sys algorithms.
}


\paragraph*{\underline{Convergence}} 
\blue{\textit{The convergence behavior of different algorithms heavily depends on the tasks; thus, it is important to support a diverse cohort in \sys.}
Figure \ref{fig:tradeoff-conv} 
illustrates the convergence behavior of
different algorithms. 
Taking VGG16 as an example and treating \sys(Allreduce) as the baseline algorithm, QSGD and Async can have almost the same convergence rate, whereas 
Decen-32bits and Decen-8bits show some drop of accuracy. 
1-bit Adam algorithm can't converge on VGG16 and the training loss explodes after a few epochs because an important assumption of 1-bit Adam regarding the gradients variance cannot hold in this task. 
For BERT-LARGE, most algorithms can converge in a similar rate as Allreduce, except Async algorithm that suffers an obvious gap.
For LSTM+AlexNet, Decen-32bits, Decen-8bits and Async can converge as Allreduce does, the performance of QSGD is degraded, and 1-bit Adam diverges again.
The above results verify that different training algorithms show diverse convergence behaviours and
there is NO algorithm that can beat others across all workloads.}
\blue{Unfortunately, given a specific task, \sys currently cannot choose the best algorithm for users.
Although a prior work~\cite{liu2021distributed} has theoretically given the convergence bounds of different training algorithms,
it is still unsolved which algorithm of them achieves the best empirical convergence rate without actually running them.
Later, we will provide some guidelines for uses based on our empirical study.}

\begin{table}[t]
\centering
\begin{tabular}{c c c c c c}
\hline
 & VGG16 & Bert-large & Bert-base & Transformer & LSTM+AlexNet \\
\hline
\sys(Async) & 48 & 57 & 421 & 185 & 128 \\
\sys(1-bit Adam) & 131 & 154 & 762 & 411 &  313 \\
\sys(QSGD) & 132 & 154 & 908 & 420 &  316 \\
\sys(Decen-32bits) & 138 & 164 & 885 & 452 & 324 \\
\sys(Decen-8bits) & 136 & 165 & 800 & 455 &  320 \\
\hline
\end{tabular}
\caption{\blue{Epoch time (s) with one straggler GPU}}
\label{tb:straggler}
\end{table}

\paragraph*{\underline{Network Conditions}} Among 
the set of algorithms that have similar convergence
behavior as Allreduce, their relative performance is
governed by the \blue{underlying network} conditions: latency
and bandwidth. We vary these two factors and 
illustrate the epoch time in Figure~\ref{fig:tradeoff-network} (We show BERT-LARGE, but other tasks have similar profile).
\blue{Algorithms that conduct communication compression 
outperforms others when the bandwidth is relatively low;
whereas decentralized algorithms outperform others
when the latency is relatively high.} 
We see 
when the network gets slower, \blue{the gap between \sys 
and other systems becomes even larger.}

\paragraph*{\underline{Worker Heterogeneity}} 
We also simulate a heterogeneous cluster \blue{by manually degrading} the \textit{Applications Clocks} of one GPU. Specifically, we set the frequency of \textit{Graphics} from 1290MHz to 585MHz. \blue{As shown in Table \ref{tb:straggler}, when there are stragglers in the system, asynchronous algorithms outperform a synchronous
one in terms of epoch time, which is also consistent
with previous observations~\cite{niu2011hogwild}.}

\paragraph*{\underline{\bf Insights and Guidlines.}}
These results justify the fundamental motivation of \sys: at the algorithmic level, there is no algorithm that can serve as a sliver bullet for all the distributed training tasks; as so, it is essential for a distributed learning system like \sys to be able to effectively fill the gap between the communication primitives defined by the infrastructure and the system relaxation demanded by various distributed learning algorithms. 
\blue{A further step beyond this empirical study is to automatically choose the optimal training algorithm for a given task and specific network condition.
However, this interesting topic is orthogonal to our work and we will explore it in the future. 
Currently, we can provide users some empirical guidelines to choose the algorithm. For example:}

\begin{itemize}
    \item 
    \blue{For the low bandwidth network, compressed algorithms (QSGD, 1-bit Adam, Decen-8bits) are likely to be better.}
    
    \item 
    \blue{For the high latency network, decentralized algorithms (Decen-32bits, Decen-8bits) are likely to be better.}
    
    \item 
    \blue{If the original optimizer is SGD, we recommend QSGD.
    While if the original optimizer is Adam, we recommend 1-bit Adam.}
    
    \item 
    \blue{If the communication/computation ratio is quite small, Async algorithm could be your choice to further reduce the communication overhead.}
\end{itemize}

\begin{table}[t]
\centering
\begin{tabular}{c c c c c c}
\hline
                & VGG16 & Bert-large & Bert-base & Transformer & AlexNet+LSTM  \\ \hline
\sys AllReduce & 105   & 114        & 510       & 318          &168        \\ 
PyTorch-DDP     & 106   & 116        & 521       &341         & 171         \\ 
Horovod         & 107   & 112        & 550       &343           & 177          \\ 
BytePS          & 170   & 114        & 548       &340            &224           \\ \hline
\end{tabular}
\caption{Epoch time (s) of the \textit{centralized full-precision synchronized} algorithm of different systems.}
\label{tab:ablation_allreduce}
\end{table}

\begin{table}[t]
\centering
\begin{tabular}{c c c c c c}
\hline
 & \makecell[c]{VGG16} & \makecell[c]{Bert-large} & \makecell[c]{Bert-base} & \makecell[c]{Transformer} & \makecell[c]{LSTM+AlexNet } \\
\hline
O=1,F=1,H=1 & 74 & 67 & 369 & 185 & 148 \\
O=0,F=1,H=1 & 88 & 70 & 395 & 185 & 163 \\
O=1,F=0,H=1 & 117 & 148 & 617 & 185 & 210 \\
O=1,F=1,H=0 & 510 & 128 & 572 & 185 & 146 \\
\hline
\end{tabular}
\caption{\blue{Epoch time (s) with different system optimizations}}
\label{tb:ablation_all}
\end{table}

\subsection{Ablation Study of System Optimizations}

We now validate the effectiveness of the \sys optimization framework. As described in Section~\ref{sys_optimization}, 
the optimization framework consists of three main optimizations:
\underline{\textbf{O}}: Overlapping between the training computation and \sys execution; \underline{\textbf{F}}: Fusion and Flattening of tensors. \underline{\textbf{H}}: Hierarchical Communications.

\blue{We first apply \sys to the standard DP-SG algorithm
and compare with PyTorch-DDP, Horovod, and BytePS, as illustrated in Table~\ref{tab:ablation_allreduce}. 
Different
from these systems that manually optimize \textit{specifically}
for DP-SG, \sys automatically optimizes for an algorithm 
that is implemented within its framework.
We see that \sys achieves similar, and sometimes better,
performance, illustrating the
effectiveness of \sys's optimization framework.}

\blue{Second, we show that all three optimizations are crucial for the end-to-end performance of \sys, and the benefits of them can vary significantly from task to task.
We conduct an ablation study and 
Table~\ref{tb:ablation_all} illustrates the result (X=0 means
the optimization X is tuned off).
We see different optimizations are important for 
different workloads.}
\blue{In principle, 
\underline{\textbf{O}} can overlap as much operations as possible,
\underline{\textbf{F}} can make communications of small tensor more efficient, and
\underline{\textbf{H}} accelerates intra-node GPU communications.}
\blue{These intuitions can be observed from the empirical results.
For communication intensive workloads (e.g., VGG-16),
hierarchical communication improves the performance significantly. 
For models with many small tensors (e.g., BERT-LARGE) and decentralized communication patterns (e.g., LSTM+AlexNet), fusion and overlapping play a larger role. An exception is the asynchronous algorithm (Transformer) because its communication is completely independent with the computation, therefore, it doesn't effect the epoch time.}

\section{Conclusion}
\label{sec:con}

We propose \sys, a communication framework whose design goal is to support various distributed training algorithms with system relaxations, powered by a new system design and a simple but effective optimization framework. We conduct
empirical study to illustrate the end-to-end performance of
\sys and to provide a systematic trandeoff study of 
different training algorithms.

\bibliographystyle{unsrt}
\bibliography{reference}

\clearpage

\appendix

\lstdefinestyle{lststyle}{
    float=tp,
    floatplacement=tbp,
    backgroundcolor=\color{backcolour},   
    commentstyle=\color{codegreen},
    keywordstyle=\color{magenta},
    numberstyle=\tiny\color{codegray},
    stringstyle=\color{codepurple},
    basicstyle=\ttfamily\scriptsize,
    breakatwhitespace=false,         
    breaklines=true,                 
    captionpos=b,                    
    keepspaces=true,                 
    numbers=left,                    
    numbersep=5pt,                  
    showspaces=false,                
    showstringspaces=false,
    showtabs=false,                  
    tabsize=1,
    frame=lines
}
\lstset{style=lststyle}
\begin{lstlisting}[language=Python, caption=End users interact with \sys 
in a familiar way,label={list:train_script}]
import torch
import bagua.torch_api as bagua
def main():
    args = parse_args()
    # define model and optimizer
    model = MyNet().to(args.device)
    optimizer = torch.optim.SGD(model.parameters(),lr=args.lr)
    # transform to BAGUA wrapper
    ######################################
    model,optimizer = bagua.bagua_init( ##
        model,                          ## <- BAGUA
        optimizer,                      ## <- WRAPPER
        bagua.algorithms.qsgd,          ##
    )                                   ##
    ######################################
    # train the model
    for epoch in range(args.epochs):
        for b_idx,(inputs,targets) in enumerate(train_loader):
            outputs = model(inputs)
            loss = torch.nn.CrossEntropyLoss(outputs,targets)
            optimizer.zero_grad()
            loss.backward()
            optimizer.step()
\end{lstlisting}

\begin{lstlisting}[language=Python, caption=Develop a Centralized Low-precision Synchronous DP-SG algorithm with error compensation in \sys,label={list:communicator}]
import bagua.torch_api as bagua

class MyAlgo():
    def __init__(self, params, optimizers, args):
        # do initialization
        self.param = params.get_flattened()
        self.optimizer = optimizers.get_flattened()
        self.args = args
        # get physical communication channel, e.g., a global communicator.
        self.global_comm = bagua.communication.get_global_comm()
        # error compensation state
        self.worker_err, self.server_err = self.global_comm.cen_lp_sync.init_states(self.param)
    def step(self):
        # get weights and gradients
        weights_flattened = self.param.data
        gradients_flattened = self.param.grad
        # execute communication with Bagua primitives, e.g., aggregating gradients over all ranks with compression
        self.global_comm.cen_lp_sync.exec(
            gradients_flattened,
            bagua.kernel.qsgd_compress_fn,
            self.worker_err,
            self.server_err
        )
        # update the weights 
        self.optimizer.step()
\end{lstlisting}

\section{Programming Interfaces of \sys}
\label{sec:appendix_program}

To help end users deploy \sys in their workloads, 
we show how to write training script and 
implement customized training algorithm.

\subsection{ML Training}
\label{sec:program_train}

Listing~\ref{list:train_script} shows a training script using \sys.
Specifically, \sys is compatible to PyTorch.
The training script contains the following steps:

\begin{itemize}
    \item{\em Define model and optimizer.} (line 6-7)
    The machine learning model is defined using PyTorch's module and placed onto a device (e.g., CPU or GPU).
    Similarly, an optimizer is defined using the \texttt{optim} module of PyTorch.
    
    \item{\em Create \sys wrapper.} (line 8-15)
    The model and optimizer are converted to \sys wrappers using the operator \texttt{bagua.bagua\_init}.
    In the \sys wrapper, the users can specify their chosen distributed training algorithm, e.g., \texttt{bagua.algorithms.qsgd} and \texttt{bagua.algorithms.allreduce}.
    With this wrapper, users can manipulate the model and optimizer in the same way they use PyTorch.
    
    \item{\em Iterative training process.} (line 16-23)
    The forward and backward computations are defined in an iterative training process.
    
\end{itemize}

As can be seen, it is easy to convert a script of PyTorch to \sys.
Users only need to convert the model and optimizer of PyTorch to \sys style using a wrapper and
the rest of the script remains the same.

\subsection{Implementation of Algorithm}
\label{sec:program_algo}

\sys provides a rich set of distributed training algorithms; however,
\sys allows developers to implement their own training algorithms.
Listing~\ref{list:communicator} shows how to implement a customized algorithm using \sys's inherited interfaces.
Typically, the users need to implement two functions --- \texttt{\_\_init\_\_()} and \texttt{step()}.

In the \texttt{\_\_init\_\_()} function, the users
1) initialize related parameters and configurations (line 5-8),
2) obtain communication channels (both global and local communicators) (line 9-10), and
3) define compression errors if data compression is applied (line 11-12).

In the \texttt{step()} function, the users can 
1) obtain the current model weights and gradients (line 14-16),
2) execute data communication between workers using the communication primitives in Bagua (e.g., \texttt{cen\_lp\_sync}), and
3) update the model weights using aggregated statistics.

\section{Overview of Distributed ML Systems}
\label{sec:appendix_related_work}

Distributed machine learning has attracted intensive 
research over the last decade, and
plenty of them are from the database community~\cite{SystemML,TeraSQLML,HybridParallelVLDB,VerticaML,DB4ML,sketchml,cost_optimizer_sigmod,jankov2019declarative,yuan2020tensor} and
machine learning community~\cite{alistarh2016qsgd,zhang2017zipml,bernstein2018signsgd,wen2017terngrad,wangni2018gradient,alistarh2018convergence,wang2018atomo,wang2017efficient,tang2019doublesqueeze,koloskova2019decentralized,li2018pipe,lian2017can,lian2018asynchronous,tang2018communication,tang2018d,wang2019adaptive,lin2019don,stich2018local,haddadpour2019local}.
\sys is built on prior research regarding distributed 
machine learning systems and algorithms.
In terms of the abstraction for communications, existing systems
fall into two paradigms: parameter server (PS) and Allreduce.

\paragraph*{\underline{Parameter Server}}

Parameter server is a distributed framework that
partitions a large model and stores them over multiple machines.
DistBelief~\cite{dean2012large} firstly proposed the framework of parameter server, with which two training algorithms, 
Downpour SGD and Sandblaster L-BFGS, were implemented. 
Li et al.~\cite{li2014scaling} designed a parameter server ML system
with a serise of optimizations, such as range push/pull, flexible consistency and user-defined filters.
Petuum~\cite{dai2015high} was proposed to support relaxed synchronization protocols.
FlexPS~\cite{FlexpsVLDB} provided the capability to change parallelism during execution.
Angel~\cite{jiang2018angel} was proposed to support hybrid paralleism and flexible server-side aggregation.
Jiang et al.~\cite{HeteroSIGMOD} studied distributed training algorithms with system heterogeneity.
PS2~\cite{PS2} integrated the parameter server architecture into a data flow ecosystem.
BytePS~\cite{byteps} presented a parameter server architecture that leveraged spare CPU and bandwidth resources in the cluster to
accelerate distributed DNN training tasks running on GPUs.
DimBoost~\cite{jiang2018dimboost} implemented a parameter server training system for gradient boosting tree models.

\begin{figure*}[t!]
	\subfigure[VGG16]{
		\scalebox{0.3}{
			\includegraphics[width=1\linewidth]{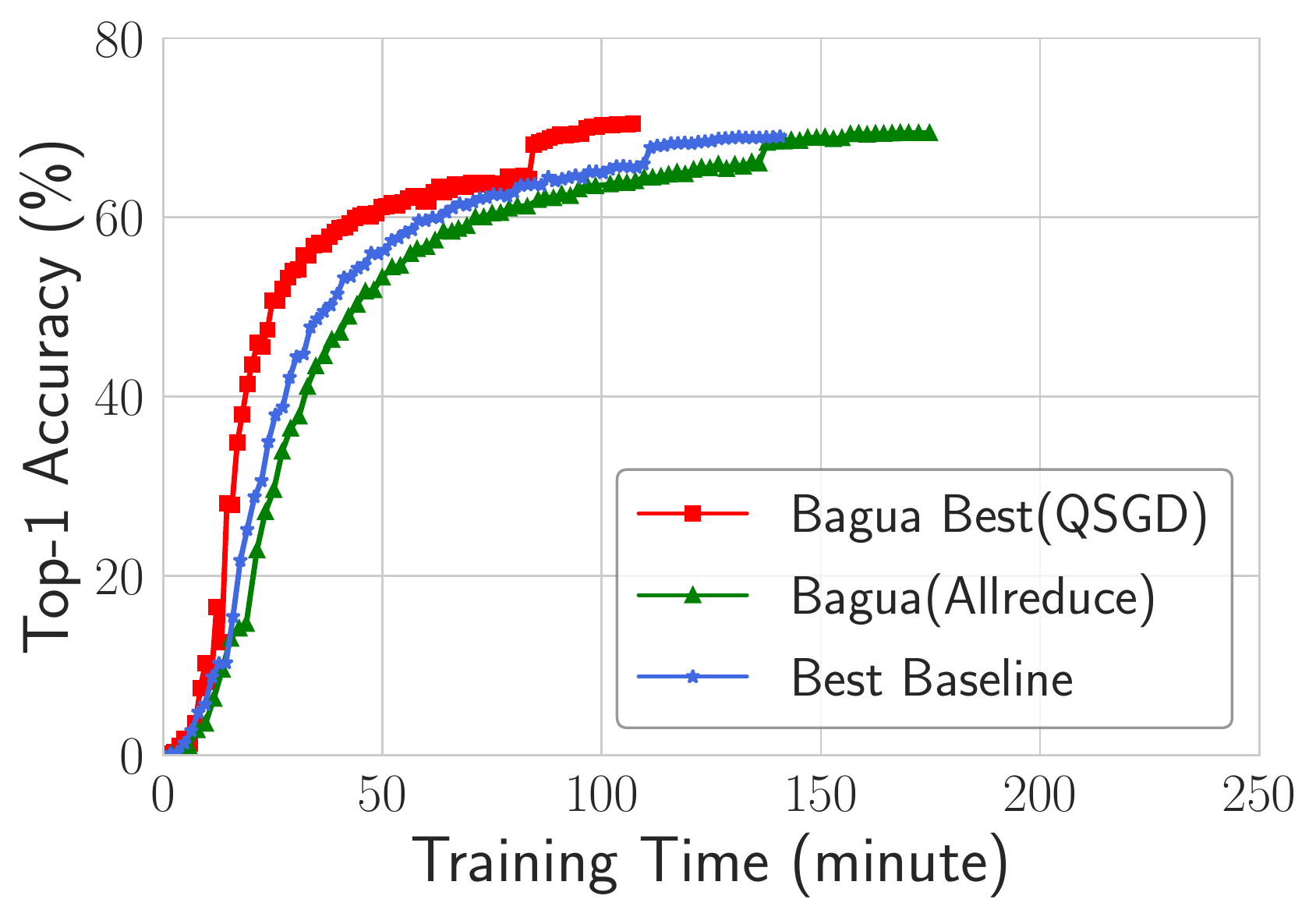}
	}}
	\subfigure[BERT-LARGE Finetune]{
		\scalebox{0.3}{
			\includegraphics[width=1\linewidth]{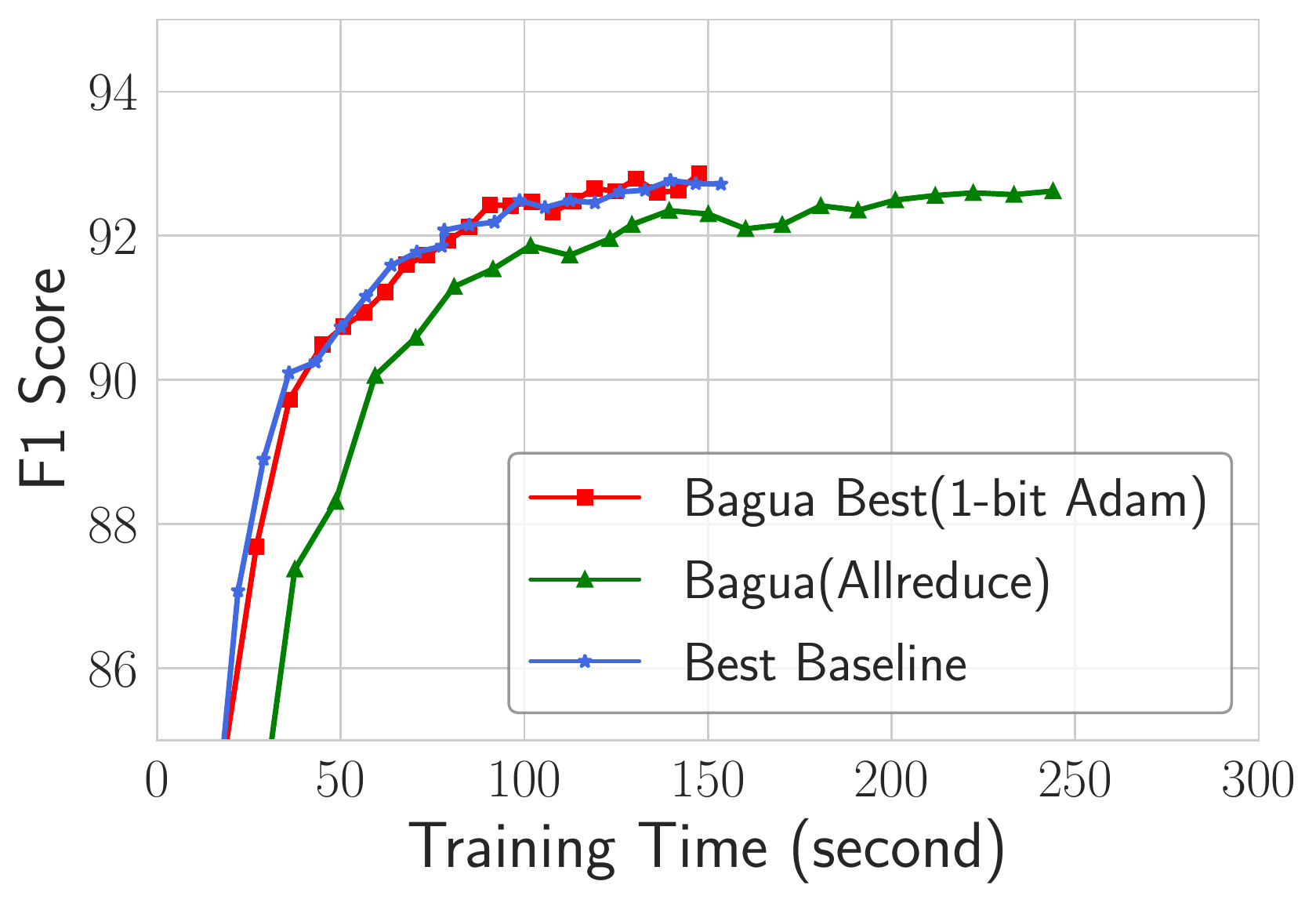}
	}}
	\subfigure[BERT-BASE Finetune]{
		\scalebox{0.3}{
			\includegraphics[width=1\linewidth]{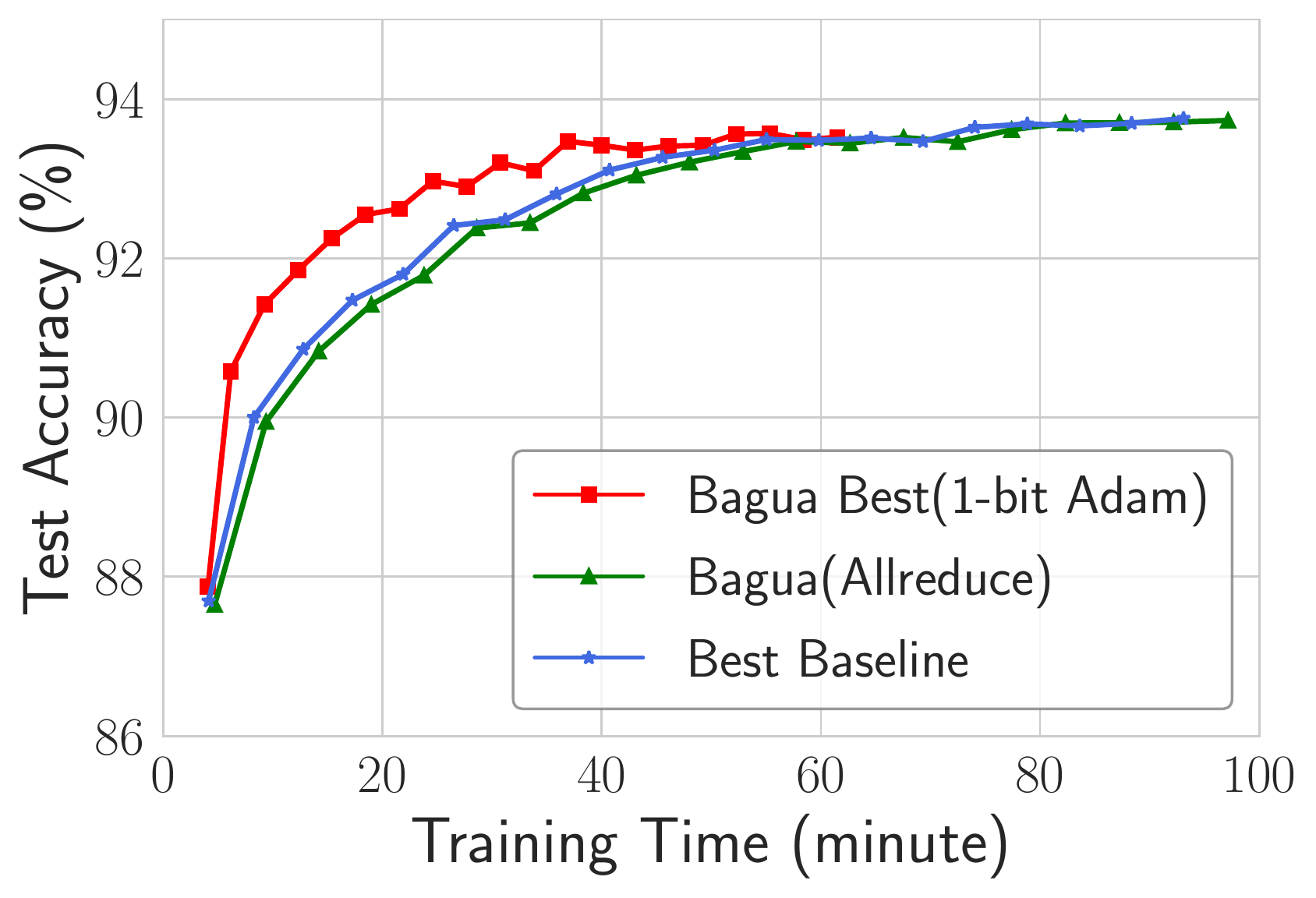}
	}}
    \subfigure[Transformer]{
		\scalebox{0.3}{
			\includegraphics[width=1\linewidth]{figures_revision/revision_e2e_transformer_10G.pdf}
	}}
    \subfigure[LSTM+AlexNet]{
		\scalebox{0.3}{
			\includegraphics[width=1\linewidth]{figures_revision/revision_e2e_hybrid_10G.pdf}
	}}
	\caption{End-to-end performance of \sys and the best competing baseline. 
	Over all five tasks, Horovod-16bits is the \textit{best} of \{Torch-DDP, Horovod-32bits, Horovod-16bits, BytePS\}. We show the performance of \sys Allreduce and the optimal \sys algorithm selected for each task. The bandwidth of inter-machine network is 100Gbps.}
  \label{fig:end2end_100G}
\end{figure*}

\paragraph*{\underline{Allreduce}}

Allreduce primitive is adopted by a range of distributed learning systems.
PyTorch Distributed~\cite{li13pytorch} extended popular PyTorch library to a distributed setting with MPI communications.
Horovod~\cite{sergeev2018horovod} provided efficient inter-GPU communication via ring reduction and 
made it easier for users to enable distributed
training in TensorFlow and PyTorch.
MLlib$^*$~\cite{zhang2019mllib} applied the technique of local update in Allreduce distributed training.
XGBoost~\cite{chen2016xgboost} leveraged Allreduce communication to train gradient boosting machines.

\section{More Experiment Results over 100Gbps Network}
\label{sec:100g}

Figure \ref{fig:end2end_100G} demonstrates the similar end-to-end experiments like Figure \ref{fig:end2end}, excepting that the network is 100Gbps here. In general, when the network is really fast, the gap between different systems and algorithms is getting smaller because the communication process takes less part of time of the entire training. For VGG16, BERT-Large and BERT-Base, although the speedup of \sys is less than it was in 10Gbps network, \sys is still the fastest one over all the competing systems. For Transformer and LSTM+AlexNet, since the active bandwidth of these two tasks is under 10Gbps, increasing bandwidth doesn't effect their training efficiency, therefore, they are able to keep the same speedup.

\end{document}